\documentclass[10pt,twocolumn,letterpaper]{article}
\usepackage[T1]{fontenc}
\usepackage{times}
\usepackage[letterpaper,textwidth=6.875in,textheight=8.875in,top=1in,columnsep=0.3125in]{geometry}
\usepackage[dvipsnames]{xcolor}
\usepackage{graphicx,amsmath,amssymb,booktabs,xspace}
\usepackage[numbers,sort&compress]{natbib}
\usepackage[format=plain,labelformat=simple,labelsep=period,font=small,compatibility=false]{caption}
\usepackage[font=footnotesize,skip=3pt,subrefformat=parens]{subcaption}

\usepackage{booktabs}   
\usepackage{array}
\usepackage{multirow}
\usepackage{xspace}
\usepackage{amsmath}
\usepackage{enumitem}
\input{generated/numbers}
\newcommand{\ADE}{ADE\xspace}

  {\par\smallskip\noindent{\color{gray}\footnotesize\textsf{OUTLINE #1}}\color{gray}\footnotesize\begin{itemize}[leftmargin=1.2em,itemsep=1pt,topsep=1pt]}%
  {\end{itemize}\normalcolor\normalsize\smallskip}

\definecolor{linkblue}{rgb}{0.21,0.49,0.74}
\usepackage[breaklinks,colorlinks,allcolors=linkblue]{hyperref}
\usepackage[capitalize]{cleveref}
\hypersetup{pdftitle={Training Lightweight Gaze Models without Target Gaze Annotation: What Makes a Good Teacher?},pdfsubject={Preprint},pdfkeywords={gaze estimation, knowledge distillation, lightweight models}}
\newcommand{\eg}{e.g.\xspace}

\date{}
\newenvironment{nolinenumbers}{}{}
\usepackage{titlesec}
\titleformat{\section}{\large\bfseries}{\thesection.}{0.5em}{}
\titleformat{\subsection}{\normalsize\bfseries}{\thesubsection.}{0.5em}{}
\titleformat{\subsubsection}{\normalsize\bfseries}{\thesubsubsection.}{0.5em}{}
\titlespacing*{\section}{0pt}{2.5ex plus 1ex minus .2ex}{1.2ex plus .2ex}
\titlespacing*{\subsection}{0pt}{2ex plus .5ex minus .2ex}{.8ex plus .2ex}

\title{Training Lightweight Gaze Models without Target Gaze Annotation:\\ What Makes a Good Teacher?}
\author{Scott J. Lewis$^{1}$ \quad Di Fu$^{2,3}$\thanks{Corresponding author: Di Fu (\href{mailto:d.fu@surrey.ac.uk}{d.fu@surrey.ac.uk}).}\\[0.6em]
{\small $^{1}$School of Computer Science and Electronic Engineering}\\
{\small $^{2}$School of Psychology}\\
{\small $^{3}$Surrey Institute for People-Centred AI (PAI)}\\
{\small University of Surrey, Guildford GU2 7XH, United Kingdom}
}
\hypersetup{pdfauthor={Scott J. Lewis; Di Fu}}
\begin{document}
\maketitle
\begin{abstract}
Model-generated labels offer a scalable route to training lightweight vision models, but what makes a model a good teacher, and can we tell before training a student? We study this in gaze target estimation, where interactive robots need models that are both accurate and efficient. In a controlled comparison of eleven supervision conditions, we vary only the source of gaze-point labels, keeping the student, training protocol and 361k training samples fixed, across three video datasets, two student sizes and three seeds; every student is evaluated against human annotation. Three findings emerge. First, imperfect model-generated labels can replace human annotation: Gaze-LLE, never trained on the target datasets' gaze annotation, trains students that match annotation-trained ones on ViT-Tiny and are significantly more accurate on ViT-Small (average distance error 0.187 vs 0.196), and the ViT-Tiny student runs 32$\times$ faster than its teacher on a laptop CPU. Second, apparent competence can conceal unreliable supervision: given another person's face centre in the prompt, InternVL3 and Qwen3-VL return it on 93.4\% and 81.1\% of test samples, and a copy-the-coordinate baseline matches InternVL3's labels and students. Without that coordinate, both models fall far behind the specialist. Third, good teachers can be identified before training: validation label error ranks student test error (Spearman's $\rho = 0.91$), and ten annotated clips single out the two specialist teachers in 95\% of draws. These findings give practical principles for assessing supervision, exposing shortcuts and selecting teachers for lightweight models.
\end{abstract}

\section{Introduction}
\label{sec:intro}
\begin{figure}[t]
\centering
\includegraphics[width=\linewidth]{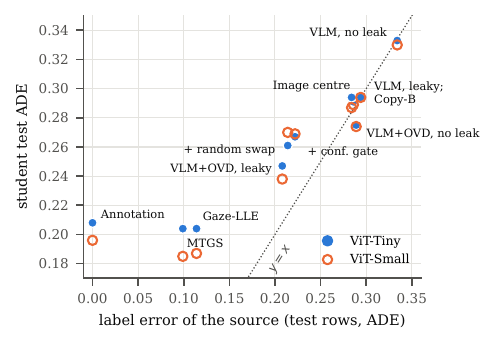}
\caption{\textbf{Label error ranks student error.} Two points per label source (ViT-Tiny filled, ViT-Small open): $x$ = the source's label average distance error (\ADE) against the annotation on the test rows, measured before training; $y$ = the student's test \ADE. The leaky VLM and copy-B points coincide. Sources: \cref{tab:sources}.}
\label{fig:labelstudent}
\end{figure}

Robots that interact with people must estimate where a person is looking, as gaze supports behaviours such as turn-taking and joint attention~\cite{admoni2017socialgaze} and anticipating a partner's intent in collaborative tasks~\cite{huang2016anticipatory}.
At the same time, robots often operate under strict computational constraints, requiring lightweight gaze models that can run locally, without relying on a network connection.
Training such models, however, typically depends on large quantities of annotated video, and obtaining frame-level gaze targets is laborious and costly.
As a result, collecting annotation can become a bottleneck in developing deployable gaze models.

Model-generated labels are commonly used in place of manual annotation~\cite{lee2013pseudolabel,xie2020noisystudent,vemulapalli2024vfmtransfer}.
For gaze target estimation, potential teachers include specialist gaze models trained on external datasets, such as Gaze-LLE~\cite{ryan2025gazelle} and MTGS~\cite{gupta2024mtgs}, vision-language models (VLMs) that can be prompted to predict gaze targets~\cite{wang2026eyesonvlm,wang2025vl4gaze}, and open-vocabulary detectors that localise named objects~\cite{liu2024groundingdino}.
It remains unclear which teacher sources produce the strongest students, and whether teacher quality can be estimated before training.
Labels can look plausible yet still mislead a student; strong agreement with a teacher's labels may not translate into agreement with the target annotation (\cref{sec:check}).
Whereas recent work evaluates VLMs directly as gaze predictors~\cite{wang2026eyesonvlm}, we focus on the effectiveness of all three as teachers for a lightweight gaze student.

We perform a controlled comparison in which the student architecture, training procedure, and \R{pop.train.all} training rows are fixed, and only the source of the coordinate labels varies (eleven sources; \cref{tab:sources}).
Experiments span three datasets (VAT, ChildPlay and VACATION), two student sizes, ViT-Tiny (\R{cost.student_tiny.params}M parameters) and ViT-Small (\R{cost.student_small.params}M), and three random seeds.
All students are evaluated against the target datasets' own gaze annotation (\R{st.tiny.gt.ntest} test rows), with significance assessed using a paired bootstrap over test clips.

We report three main findings.
First, the specialist gaze model Gaze-LLE, despite never being trained on the target datasets' gaze annotation, produces students that match or exceed those trained on the datasets' own annotation, with significantly lower error on ViT-Small (\R{st.small.gazelle.pooled} vs \R{st.small.gt.pooled}; \cref{sec:teacher}).
Because Gaze-LLE requires \R{cost.gazelle_vitb14.flops_x_tiny}$\times$ the ViT-Tiny student's compute, it is used once, offline, to generate training labels.
Second, prompt leakage can create a false impression of gaze understanding: when its prompt includes another person's face centre, InternVL3 reproduces that coordinate in \R{copy.test.all} of its test labels, and Qwen3-VL behaves the same way.
A baseline that simply returns the supplied coordinate exposes this behaviour, and removing the leak reveals a substantial gap between general-purpose VLMs and specialist gaze models (\cref{sec:vlm}).
Third, a teacher's label error, measured in seconds before any training, closely ranks student performance (Spearman's $\rho = \R{lcv.rho.tiny}$; \cref{sec:check}), and ten annotated clips are enough to identify the best teachers.
The results also show that confidence gating is no better than random replacement, and that apparent temporal gains come almost entirely from biased teachers rather than anticipation (\cref{sec:check}).

In summary, our contributions are:
\begin{itemize}
\item We show that a specialist gaze model never trained on the target datasets' gaze annotation can replace it when training lightweight gaze students, at both sizes, with or without head-position input.
\item We identify a hidden pitfall in VLM-based gaze labelling: prompt leakage can create the appearance of competence while the model largely reproduces the supplied coordinate, which a simple copy-the-coordinate baseline exposes.
\item We show that teacher quality can be estimated before student training from label error measured on only a few annotated clips and give three practical recommendations for evaluating model-generated supervision.
\end{itemize}
We will release the label sets, code and evaluation protocol.

\section{Related work}
\label{sec:related}

\paragraph{Gaze target estimation.}
Gaze target estimation predicts where a person in an image is looking~\cite{recasens2015gazefollow}.
Prior work has extended this setting to video and out-of-frame gaze targets~\cite{chong2018connecting,chong2020vat} and to children~\cite{tafasca2023childplay}.
Methods include transformers for multi-person gaze target estimation~\cite{tu2022hgttr,tafasca2024sharingan}, temporal and social-gaze models~\cite{gupta2024mtgs}, and light decoders on frozen foundation encoders~\cite{ryan2025gazelle}.
Recent work has improved gaze reasoning to reduce reliance on salient objects~\cite{wang2026gazereasoning} or introduced promptable gaze target estimation, where textual or visual prompts specify the person of interest~\cite{cao2026gazeanywhere}.
Research has also explored social attention in robotics~\cite{fu2023humanoid,abawi2025hri}.
Unlike these gaze models, which are typically trained on the target benchmarks, we compare sources of model-generated supervision for lightweight gaze students and judge the students against the target annotation.

\paragraph{Vision-language models and gaze.}
VLMs have been used to extract contextual cues that improve gaze-following models~\cite{gupta2024vlmgaze} and have been fine-tuned or prompted to describe and locate gaze targets~\cite{wang2025vl4gaze,miao2026omnigf}.
A controlled study reports that 94 of the 111 models tested fail to do better than random guessing at identifying the object a person is looking at, while humans achieve near-ceiling accuracy~\cite{zhang2025vlmgaze}.
A benchmark of VLMs as gaze predictors finds that zero-shot Qwen3-VL-8B has more than twice the error of Gaze-LLE on GazeFollow~\cite{wang2026eyesonvlm}.
Whereas these studies evaluate or improve VLMs as gaze predictors, we evaluate them as sources of supervision for lightweight students.
We further identify a shortcut not reported in these works: a prompt that contains a candidate answer lets a VLM appear competent by copying it, a behaviour exposed by a copy-the-coordinate baseline.

\paragraph{Distillation and model-generated labels.}
Knowledge distillation transfers knowledge from a teacher model to a smaller student~\cite{hinton2015distilling,gou2021kdsurvey}.
Students can match or exceed a teacher of the same size~\cite{furlanello2018bornagain}, while teacher-student capacity gaps, optimisation and training duration influence outcomes~\cite{cho2019efficacy,mirzadeh2020teacherassistant,stanton2021kdwork,beyer2022patient}.
Model-generated supervision is also used through pseudo-labelling, self-training and transfer from vision foundation models~\cite{lee2013pseudolabel,xie2020noisystudent,vemulapalli2024vfmtransfer}.
In gaze target estimation, PaGE trains a DINOv3-based teacher on the labelled training sets of GazeFollow, VAT and ChildPlay, reported to reach human-level accuracy, then distils lightweight students on unlabelled images (MPII, OpenImages~V7) and fine-tunes them on the same labelled training sets~\cite{ye2026page}.
Distillation has also been applied to near-eye eye tracking for AR/VR~\cite{jiang2026distillgaze} and to robots for social navigation~\cite{elnoor2025vilam} and proactive interaction~\cite{desimone2025proactive}.
Unlike PaGE, which uses target-dataset annotation to train both teacher and students, we treat that annotation as one competing source of supervision, evaluate whether existing models can replace it, and score each teacher's labels against it before student training.

\paragraph{Noisy labels and evaluation.}
A substantial body of work studies learning under noisy supervision~\cite{song2023noisysurvey}: deep networks can memorise even random labels~\cite{zhang2017rethinking} but tend to learn clean patterns first~\cite{arpit2017memorization,liu2020elr}.
Softened targets such as label smoothing prevent over-confidence~\cite{szegedy2016inception,muller2019labelsmoothing}.
Other methods select likely-clean samples by their loss~\cite{han2018coteaching}, identify label errors from predicted probabilities~\cite{northcutt2021confident}, or weight teachers by their confidence against ground truth~\cite{zhang2022confidenceaware}.
Evaluation protocols can also change conclusions, through benchmark variance~\cite{bouthillier2021variance} or metrics that suggest a success that does not hold, as in machine unlearning~\cite{george2025illusion}.
Whereas noisy-label methods seek to make training robust, we measure label error against the annotation before training, show that confidence-guided replacement is no better than random replacement at a matched rate, and derive recommendations for evaluating model-generated supervision.

\section{Experimental setup}
\label{sec:setup}
All comparisons keep the student, the data, the training recipe and seeds fixed, differing only in where the coordinate targets come from.

\begin{table*}[t]
\centering
\small
\setlength{\tabcolsep}{4pt}
\begin{tabular}{@{}l >{\raggedright\arraybackslash}p{4.4cm} >{\raggedright\arraybackslash}p{6.2cm} >{\raggedright\arraybackslash}p{3.4cm}@{}}
\toprule
Source & Model & How the label is produced (social rows\,/\,object rows) & Trained on target datasets? \\
\midrule
\C{gt} & annotation & annotated point & n/a \\
\C{mtgs} & MTGS~\cite{gupta2024mtgs} & model prediction from the video clip & VAT, ChildPlay (VSGaze); not VACATION \\
\C{gazelle} & Gaze-LLE~\cite{ryan2025gazelle}: frozen DINOv2 ViT-B/14~\cite{oquab2024dinov2} + decoder & arg-max of the $64\times64$ heatmap & no (GazeFollow~\cite{recasens2015gazefollow} only) \\
\midrule
\C{teacher_internvl3_alone} & InternVL3-8B~\cite{zhu2025internvl3} & answer to a prompt that also gives person B's head box and face centre & unknown (web-scale data) \\
\C{copyb} & none & the face centre supplied in that prompt & n/a \\
\C{vlmnoleak} & InternVL3-8B & answer to a prompt with person A's head box only & unknown (web-scale data) \\
\C{teacher_hybrid} & InternVL3-8B + Grounding DINO~\cite{liu2024groundingdino} & leaky VLM\,/\,centre of the detector's best box for the VLM's object phrase; annotation when no box passes the thresholds & unknown (web-scale data) \\
\C{ovdalwaysdino} & InternVL3-8B + Grounding DINO & no-leak VLM\,/\,best box on every object row, no thresholds & unknown (web-scale data) \\
\C{ovdgate55} & as above & as above, object labels with detector score $<0.55$ replaced by the annotation & as above \\
\C{ovdmixrand} & as above & as above, the same share replaced at random & as above \\
\midrule
\C{centre} & none & $(0.5,\,0.5)$ for every row & n/a \\
\bottomrule
\end{tabular}
\caption{\textbf{Supervision sources.} Every source labels the same rows, with the same student, recipe and seeds. Person A is the person whose gaze is labelled; person B is another annotated person in the frame. The last column shows whether a labeller may have seen the target datasets in training.}
\label{tab:sources}
\end{table*}

\paragraph{Data and population.}
We use three datasets: VideoAttentionTarget (VAT)~\cite{chong2020vat}, ChildPlay~\cite{tafasca2023childplay} and VACATION~\cite{fan2019vacation}.
All provide per-person gaze-target annotations.
For VAT and ChildPlay, the gaze type is derived from the annotation: \emph{social}, if the gaze point lies inside another person's head box; \emph{off-screen}, if the annotation places the target outside the frame; otherwise \emph{object}.
VACATION's annotations contain no gaze points; instead its `attention-focus' field states the person or object being looked at.
We use the target bounding-box centre as ground truth (social if it is a person's face, object otherwise).
Rows (one person in one frame) with ambiguous or undefined targets are dropped, yielding \R{st.tiny.gt.ntest} on-screen test rows (VAT \R{st.tiny.gt.ntest.vat}; ChildPlay \R{st.tiny.gt.ntest.childplay}; VACATION \R{st.tiny.gt.ntest.vacation}) and \R{io.test.pos} off-screen test rows (VAT only).
We apply their canonical split for all datasets except VAT's validation set, where we use the split MTGS uses~\cite{tafasca2023childplay,gupta2024mtgs}, ensuring no frame or clip appears in more than one partition.

\paragraph{Supervision sources.}
\Cref{tab:sources} compares eleven supervision sources.
\emph{Specialist gaze models.}
MTGS~\cite{gupta2024mtgs} models the gaze of all subjects jointly.
The released checkpoint with no fine-tuning is used.
Gaze-LLE~\cite{ryan2025gazelle} decodes from a frozen DINOv2 ViT-B/14 backbone~\cite{oquab2024dinov2}.
The GazeFollow-only~\cite{recasens2015gazefollow} checkpoint is used, since the in/out variants are fine-tuned on VAT and ChildPlay.

\emph{Vision-language models.}
InternVL3-8B~\cite{zhu2025internvl3} with greedy decoding is tested with two prompts.
\emph{No-leak}: the model is shown the frame and person A's head box and asked for the point A is looking at: a face centre if the target is a person, an object centre otherwise.
\emph{Leaky}: the model is also given the head box and face centre of another annotated person in the scene, `person B' (chosen without reference to the gaze annotation).
The prompt asks the model to return that centre if A is looking at B.
\emph{Copy-B} always returns the supplied coordinate, providing the results the leaky VLM would produce if it only copied (prompts: supplementary).
Qwen3-VL-8B~\cite{bai2025qwen3vl} is run with the same two prompts and evaluated at label level only (\cref{sec:vlm}).

\emph{Detector-assisted labels.}
VLM+OVD pairs InternVL3-8B with an open-vocabulary detector (OVD), Grounding DINO (base)~\cite{liu2024groundingdino}.
Social rows default to the VLM's answer; for object rows, the VLM names the target from the frame alone, without being told whose gaze to estimate, and Grounding DINO's highest-scoring box for that target is used.
While the no-leak version always accepts this box, the leaky version retains the detector's thresholds, falling back to the annotation if no box passes them or the VLM names no object.
Two controls test confidence gating: \emph{gated} replaces a detector label with the annotation if its score falls below 0.55, whereas \emph{random swap} replaces an identical number of labels but selects the rows at random.
This comparison is repeated at thresholds 0.40, 0.48 and 1 (supplementary).
The image-centre baseline always predicts the centre of the image.
Whilst every source uses annotated head boxes and off-screen flags, teachers only use annotation in four situations: VLM+OVD's routing by gaze type; VAT frames without a person B (leaky VLM and copy-B; \R{unmatched.share.vat} of VAT training rows); leaky VLM+OVD's fall-backs (\R{lab.train.all.ovd_old.all.gtshare1} of training rows); and the gating controls.

\paragraph{Student.}
Two students, ViT-Tiny (\R{cost.student_tiny.params}M parameters) and ViT-Small (\R{cost.student_small.params}M parameters)~\cite{dosovitskiy2021vit} (patch size 16, ImageNet-pretrained weights from timm~\cite{wightman2019timm}), take two inputs: the full image (resized to $224\times224$) and a crop of the subject's head ($112\times112$ upsampled to $224\times224$).
The class tokens are concatenated and passed through a two-layer MLP (LayerNorm, GELU, dropout), to a sigmoid output (\emph{gaze point}) and a logit for off-screen prediction.
Each student's input includes neither a head-position map nor any other person's box.
A variant additionally receives head position (\cref{sec:teacher}).

\paragraph{Training.}
For on-screen targets, the coordinate loss is Smooth-$\ell_1$.
Off-screen output uses binary cross-entropy (weight 1).
For every condition, the off-screen flag comes from the annotation.
Training uses AdamW~\cite{loshchilov2019adamw} (learning rate $3\times10^{-5}$, weight decay $1\times10^{-4}$), batch size 32, and cosine decay~\cite{loshchilov2017sgdr} to 1\% of the initial rate over 15 epochs.
All conditions are trained with three seeds: 42, 88 and 137 (full recipe: supplementary).

\paragraph{Selection and scoring.}
We select each student's epoch using its validation error against \emph{its own training labels} (selection on ground truth: supplementary).
Students are then scored against the annotated gaze point, not the teacher's prediction.
We report the Average Distance Error (\ADE) between predicted and annotated points, in image coordinates normalised to $[0,1]$, as mean\,$\pm$\,standard deviation over the three seeds, and the Percentage of Correct Keypoints at 0.1 (PCK$_{.1}$), the fraction of predictions within 0.1 of the annotated point (\cref{sec:check}).
We then test the differences between sources with a paired bootstrap~\cite{field2007clusterbootstrap} over \R{sig.tiny.gazelle.vs.gt.all.clips} test clips on seed-averaged per-row errors (10{,}000 resamples, 95\% intervals).
Differences are computed from unrounded errors, so they can differ by 0.001 from the rounded values in the tables.
For the temporal heads (\cref{sec:check}), the architecture is chosen on a held-out fold of the test clips, after which the chosen head is evaluated on a separate fold (five-fold cross-fitting).
The off-screen decision threshold is chosen on validation (maximum F1), then applied unchanged to test (results: \cref{sec:check}).

\paragraph{What each labeller may have seen.}
MTGS was trained on VSGaze~\cite{gupta2024mtgs}, which incorporates VAT and ChildPlay.
Consequently, labels generated on these training rows may partly reproduce their annotation.
Since our partitions match MTGS's splits, none of our test rows are in its training data.
VACATION is unseen by MTGS, allowing it to be used as a held-out check.
Gaze-LLE has never been trained on the target datasets' gaze annotation: its decoder was fitted only on GazeFollow, a still-image dataset, and its DINOv2 backbone is frozen.
The training data of DINOv2, the VLMs and Grounding DINO is web-scale and not fully disclosed, making overlap with the target datasets difficult to assess.

\section{A specialist teacher can replace target gaze-point annotation}
\label{sec:teacher}
Gaze-LLE was never trained on the target datasets' gaze annotation, yet students trained on its labels match or exceed those trained on that annotation (\cref{tab:main}).
ViT-Tiny students are comparable (\ADE \R{st.tiny.gazelle.pooled} vs \R{st.tiny.gt.pooled}), whilst ViT-Small students are significantly more accurate (\ADE \R{st.small.gazelle.pooled} vs \R{st.small.gt.pooled}), with \R{sig.small.gazelle.vs.gt.all.seedpairs.same} seed pairings favouring Gaze-LLE.

\paragraph{MTGS.}
MTGS mirrors Gaze-LLE's results for both students (\ADE \R{st.tiny.mtgs.pooled} on ViT-Tiny, \R{st.small.mtgs.pooled} on ViT-Small).
Because it was trained on VAT and ChildPlay, VACATION is its held-out check, where it matches the annotation on ViT-Tiny (\R{st.tiny.mtgs.vacation} vs \R{st.tiny.gt.vacation}; \cref{tab:perdataset}).

\paragraph{Why imperfect labels can be effective supervision.}
Despite differing from the human gaze-point annotation (test label error \R{lab.test.all.mtgs.all} for MTGS and \R{lab.test.all.gazelle.all} for Gaze-LLE), specialist teachers produce students that perform comparably to, and in some configurations better than, those trained on the annotation.
Gaze-LLE's labels are easier to fit: students trained on Gaze-LLE reproduce its labels more closely than annotation-trained students reproduce the annotation (0.191 vs \R{st.tiny.gt.pooled} on ViT-Tiny, significant for both student sizes).
A teacher's effectiveness may therefore depend not only on its agreement with the annotation, but also on how easily a student can fit its labels.
\Cref{fig:labelstudent} shows a pattern consistent with this possibility: differences in label error between relatively accurate teachers translate into comparatively small differences in student performance, whereas larger teacher errors are more strongly reflected in performance.

\paragraph{Adding head position.}
Students trained on teacher and annotation labels may appear comparable only because the student is the bottleneck: it sees the head crop, but not the head's location within the image.
To test this, we retrained ViT-Tiny students on three sources (annotation, Gaze-LLE, no-leak VLM) and ViT-Small students on two (annotation, Gaze-LLE), adding the head's position in the frame as an input (\cref{tab:headpos}).
This significantly improves the ViT-Tiny students trained on the annotation (\R{sig.headposgain.tiny.gt.diff}) and on Gaze-LLE (\R{sig.headposgain.tiny.gazelle.diff}), but not those trained on the no-leak VLM (\R{sig.headposgain.tiny.vlmnoleak.diff}, \R{sig.headposgain.tiny.vlmnoleak.sig}): a better-informed student cannot compensate for inaccurate labels.
For ViT-Tiny with head position, Gaze-LLE's students remain comparable to the annotation's (\ADE \R{headpos.tiny.gazelle.pooled} vs \R{headpos.tiny.gt.pooled}).
For ViT-Small, the added input helps the annotation-trained students more than the Gaze-LLE-trained ones (\R{sig.headposgain.small.gt.diff} vs \R{sig.headposgain.small.gazelle.diff}, both significant), narrowing Gaze-LLE's advantage to a difference that is not significant (\ADE \R{headpos.small.gazelle.pooled} vs \R{headpos.small.gt.pooled}).
Gaze-LLE therefore matches the annotation even with a better-informed student.

\paragraph{Accuracy and cost.}
Gaze-LLE is more accurate than any student model (test label error \R{lab.test.all.gazelle.all}), but it needs \R{cost.gazelle_vitb14.flops_x_tiny}$\times$ the ViT-Tiny student's compute, which makes direct deployment impractical without a GPU (\cref{tab:cost}).
On a 2021 laptop, Gaze-LLE is \R{cost.gazelle_vitb14.cpu4_x_tiny}$\times$ slower than ViT-Tiny on the CPU (\R{cost.gazelle_vitb14.cpu4_ms} vs \R{cost.student_tiny.cpu4_ms}\,ms per frame) and \R{cost.gazelle_vitb14.gpu_x_tiny}$\times$ slower on the GPU.
The most practical approach is to use Gaze-LLE only once, offline, to generate training targets, while the student model provides the efficiency required for deployment to a robot.

\begin{table*}[t]
\centering
\footnotesize
\setlength{\tabcolsep}{4pt}
\begin{tabular}{@{}l ccc c cccc c cc@{}}
\toprule
 & \multicolumn{3}{c}{Label error (test rows)} & & \multicolumn{4}{c}{ViT-Tiny student} & & \multicolumn{2}{c}{ViT-Small student} \\
\cmidrule(lr){2-4}\cmidrule(lr){6-9}\cmidrule(l){11-12}
Supervision & \ADE & social & object & & \ADE & social & object & PCK$_{.1}$ & & \ADE & PCK$_{.1}$ \\
\midrule
\C{gt} & 0 & 0 & 0 &  & 0.208\,{\scriptsize$\pm$0.003} & 0.178 & 0.233 & 0.201 &  & 0.196\,{\scriptsize$\pm$0.002} & 0.243 \\
\C{mtgs} & 0.099 & 0.082 & 0.116 &  & 0.204\,{\scriptsize$\pm$0.004} & 0.168 & 0.234 & 0.215 &  & 0.185\,{\scriptsize$\pm$0.002} & 0.282 \\
\C{gazelle} & 0.114 & 0.094 & 0.131 &  & 0.204\,{\scriptsize$\pm$0.000} & 0.171 & 0.232 & 0.213 &  & 0.187\,{\scriptsize$\pm$0.004} & 0.276 \\
\C{teacher_hybrid} & 0.208 & 0.151 & 0.256 &  & 0.247\,{\scriptsize$\pm$0.003} & 0.209 & 0.279 & 0.131 &  & 0.238\,{\scriptsize$\pm$0.006} & 0.155 \\
\C{teacher_internvl3_alone} & 0.286 & 0.149 & 0.403 &  & 0.291\,{\scriptsize$\pm$0.001} & 0.168 & 0.397 & 0.163 &  & 0.289\,{\scriptsize$\pm$0.005} & 0.212 \\
\C{copyb} & 0.284 & 0.138 & 0.409 &  & 0.294\,{\scriptsize$\pm$0.005} & 0.163 & 0.406 & 0.199 &  & 0.287\,{\scriptsize$\pm$0.000} & 0.241 \\
\C{vlmnoleak} & 0.334 & 0.325 & 0.342 &  & 0.333\,{\scriptsize$\pm$0.007} & 0.287 & 0.372 & 0.052 &  & 0.330\,{\scriptsize$\pm$0.002} & 0.042 \\
\addlinespace[2pt]
\C{ovdalwaysdino} & 0.289 & 0.325 & 0.259 &  & 0.275\,{\scriptsize$\pm$0.004} & 0.259 & 0.289 & 0.076 &  & 0.274\,{\scriptsize$\pm$0.017} & 0.073 \\
\C{ovdgate55} & 0.222 & 0.325 & 0.134 &  & 0.267\,{\scriptsize$\pm$0.003} & 0.258 & 0.276 & 0.066 &  & 0.269\,{\scriptsize$\pm$0.013} & 0.075 \\
\C{ovdmixrand} & 0.214 & 0.325 & 0.118 &  & 0.261\,{\scriptsize$\pm$0.005} & 0.252 & 0.269 & 0.069 &  & 0.270\,{\scriptsize$\pm$0.005} & 0.085 \\
\midrule
\C{centre} & 0.294 & 0.284 & 0.302 &  & 0.294\,{\scriptsize$\pm$0.000} & 0.284 & 0.302 & 0.061 &  & 0.294 & 0.061 \\
\bottomrule
\end{tabular}

\caption{\textbf{Students of every source, scored against the annotation.} Label \ADE (test rows) beside student test \ADE (mean\,$\pm$\,sd, 3 seeds) and PCK$_{.1}$ (share of predictions within 0.1 of the annotated point; higher is better). Epochs are selected on each source's own validation labels. The image-centre `student' is a constant prediction.}
\label{tab:main}
\end{table*}
\begin{table}[t]
\centering
\scriptsize
\setlength{\tabcolsep}{2.4pt}
\begin{tabular}{@{}l ccc ccc@{}}
\toprule
 & \multicolumn{3}{c}{ViT-Tiny} & \multicolumn{3}{c}{ViT-Small} \\
\cmidrule(lr){2-4}\cmidrule(l){5-7}
Supervision & VAT & CP & VAC & VAT & CP & VAC \\
\midrule
\C{gt} & 0.215 & 0.192 & 0.223 & 0.208 & 0.180 & 0.201 \\
\C{mtgs} & 0.208 & 0.192 & 0.216 & 0.192 & 0.174 & 0.193 \\
\C{gazelle} & 0.209 & 0.191 & 0.217 & 0.197 & 0.175 & 0.192 \\
\midrule
MTGS $-$ annot. & $-$0.006 & $-$0.001 & $-$0.007 & $-$0.016$^{*}$ & $-$0.006 & $-$0.008 \\
Gaze-LLE $-$ annot. & $-$0.006 & $-$0.001 & $-$0.006 & $-$0.011$^{*}$ & $-$0.005 & $-$0.009 \\
\bottomrule
\end{tabular}

\caption{\textbf{Specialist teachers by dataset.} Per-dataset student test \ADE (mean of 3 seeds; CP = ChildPlay, VAC = VACATION). Bottom rows: teacher minus annotation, paired clip bootstrap; $^{*}$ = significant. VACATION is held out for both teachers; Gaze-LLE was never trained on the gaze annotation of any of the three datasets.}
\label{tab:perdataset}
\end{table}
\begin{table}[t]
\centering\footnotesize\setlength{\tabcolsep}{2.3pt}
\begin{tabular}{@{}l cc cc@{}}
\toprule
 & \multicolumn{2}{c}{without head position} & \multicolumn{2}{c}{with head position} \\
\cmidrule(lr){2-3}\cmidrule(l){4-5}
Supervision & \ADE $\downarrow$ & PCK$_{.1}\uparrow$ & \ADE $\downarrow$ & PCK$_{.1}\uparrow$ \\
\midrule
\multicolumn{5}{@{}l}{\textit{ViT-Tiny}} \\
\C{gt} & 0.208\,{\scriptsize$\pm$0.003} & 0.201 & 0.199\,{\scriptsize$\pm$0.002} & 0.218 \\
\C{gazelle} & 0.204\,{\scriptsize$\pm$0.000} & 0.213 & 0.193\,{\scriptsize$\pm$0.005} & 0.252 \\
\C{vlmnoleak} & 0.333\,{\scriptsize$\pm$0.007} & 0.052 & 0.333\,{\scriptsize$\pm$0.003} & 0.047 \\
Gaze-LLE $-$ annotation & $-$0.004 & & $-$0.005 & \\
\quad 95\% CI & \multicolumn{2}{l}{[$-$0.009, +0.001]} & \multicolumn{2}{l}{[$-$0.011, +0.001]} \\
\midrule
\multicolumn{5}{@{}l}{\textit{ViT-Small}} \\
\C{gt} & 0.196\,{\scriptsize$\pm$0.002} & 0.243 & 0.186\,{\scriptsize$\pm$0.004} & 0.273 \\
\C{gazelle} & 0.187\,{\scriptsize$\pm$0.004} & 0.276 & 0.183\,{\scriptsize$\pm$0.004} & 0.292 \\
Gaze-LLE $-$ annotation & $-$0.008 & & $-$0.003 & \\
\quad 95\% CI & \multicolumn{2}{l}{[$-$0.013, $-$0.004]} & \multicolumn{2}{l}{[$-$0.008, +0.001]} \\
\bottomrule
\end{tabular}

\caption{\textbf{Adding head position} (3 seeds). Student test \ADE and PCK$_{.1}$ without and with the head's position in the frame. `Gaze-LLE $-$ annotation' = difference in pooled test \ADE (negative = Gaze-LLE better; computed before rounding), with the 95\% CI from the paired clip bootstrap. The no-leak VLM control is run on ViT-Tiny only.}
\label{tab:headpos}
\end{table}
\begin{table}[t]
\centering\footnotesize\setlength{\tabcolsep}{2pt}
\begin{tabular}{@{}l r r r r r c@{}}
\toprule
 & Params & & \multicolumn{2}{c}{ms/frame} & GPU mem. & Test \\
\cmidrule(lr){4-5}
Model & (M) & GFLOPs & CPU & GPU & (MB) & \ADE $\downarrow$ \\
\midrule
ViT-Tiny student & 5.7 & 5.0 & 34 & 12.1 & 31 & 0.204 \\
ViT-Small student & 21.9 & 18.4 & 98 & 12.4 & 93 & 0.187 \\
\midrule
Gaze-LLE ViT-B/14 & 89.4 & 222.8 & 1{,}077 & 62.7 & 1{,}993 & 0.114 \\
Gaze-LLE ViT-L/14 & 307.3 & 732.7 & 4{,}299 & 190.9 & 3{,}371 & n/a \\
\bottomrule
\end{tabular}

\caption{\textbf{Accuracy and cost per frame} (one person, batch 1, fp32). Laptop \R{cost.cpu} CPU (4 threads) and \R{cost.gpu} GPU; GFLOPs counted with PyTorch's FLOP counter, including the DINOv2 backbone. Student \ADE: Gaze-LLE-trained students; Gaze-LLE \ADE: its own test label error. ViT-L/14 is shown for scale (not used as a teacher).}
\label{tab:cost}
\end{table}

\section{When the prompt contains the answer}
\label{sec:vlm}
\begin{figure*}[t]
\centering
\includegraphics[width=\linewidth]{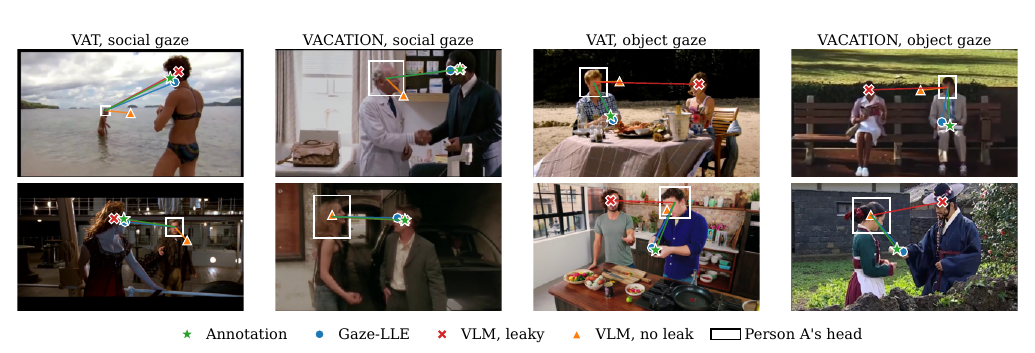}
\caption{\textbf{Labels from three teachers on typical test frames.} Eight test frames from VAT and VACATION, chosen so that every source's label error is close to its median for that gaze type (not best cases); lines run from person A's head to each label. The leaky VLM's label lands on person B's face, the coordinate in its prompt: correct on social gaze, wrong on object gaze. Gaze-LLE is close in all eight.}
\label{fig:qualitative}
\end{figure*}
When given the leaky prompt, which includes another person's face centre, a VLM appears to be competent on social gaze, but it is mostly coordinate copying rather than gaze reasoning (\cref{fig:qualitative}).
InternVL3 reproduces that point almost exactly (to within 0.001) on \R{copy.test.all} of test rows with a person B, ranging from \R{copy.test.childplay} on ChildPlay to \R{copy.test.vat} on VAT.
Copy-B (\cref{sec:setup}), which always returns the supplied coordinate, exposes this behaviour: both the generated labels and the resulting students match those obtained from the VLM.
Label error is nearly identical (\ADE \R{lab.test.all.vlmleak_old.all} vs \R{lab.test.all.copyb.all}), and students trained on these labels are not significantly different at either size (ViT-Tiny \R{st.tiny.teacher_internvl3_alone.pooled} vs \R{st.tiny.copyb.pooled}).
These students are about as accurate as always predicting the image centre (\R{st.tiny.centre.pooled}).

\paragraph{Copying scores well on social rows.}
On social rows, the leaky labels look accurate (label error \R{lab.test.all.vlmleak_old.social}, copy-B \R{lab.test.all.copyb.social}) because the supplied face centre is usually the answer.
When A looks at another person, that person is often B, and in two-person scenes it can only be B.
On object rows, where the supplied face centre is never the answer, the leaky labels are far worse than the image centre (\R{lab.test.all.vlmleak_old.object} vs \R{lab.test.all.centre.object}).

\paragraph{Both tested models copy.}
With the same leaky prompt, Qwen3-VL-8B reproduces the supplied face centre on \R{svlm.pooled.qwen3vl8b_leaky.copy} of the same rows, ranging from \R{svlm.childplay.qwen3vl8b_leaky.copy} on ChildPlay to \R{svlm.qwen3vl8b_leaky.copy} on VAT (InternVL3: \R{svlm.pooled.internvl3_leaky.copy}; \cref{tab:vlm}).
Copying is therefore the dominant behaviour for both models, but its prevalence depends on the model and data, which is why the copy-B check should be run for each.

\paragraph{Without the leak.}
InternVL3 produces labels that are less accurate than predicting the image centre on every dataset and both gaze types (label error \R{svlm.pooled.internvl3_noleak.all} vs \R{lab.test.all.centre.all}), and students trained on these labels also underperform the centre at both sizes (ViT-Tiny \R{st.tiny.vlmnoleak.pooled} vs \R{st.tiny.centre.pooled}).
In contrast, Qwen3-VL surpasses the image-centre baseline (\R{svlm.pooled.qwen3vl8b_noleak.all} vs \R{lab.test.all.centre.all}), indicating that it extracts some gaze information directly from the image.
However, its label error remains about twice that of Gaze-LLE (\R{lab.test.all.gazelle.all}).
Thus, without the leak, both general-purpose VLMs remain substantially less accurate than a specialist gaze model.

\paragraph{Distinguishing visual inference from prompt-induced shortcuts.}
The apparent accuracy of VLM-generated labels can partly arise from information supplied in the prompt rather than independently inferred from visual evidence.
This distinction is particularly important when VLMs are used as teachers, as downstream students may inherit systematic shortcuts embedded in their supervision.
Comparing VLM outputs against a simple baseline that directly reproduces the supplied coordinates therefore provides a practical diagnostic for shortcut exploitation.
More broadly, predictive accuracy alone may be insufficient to establish the reliability of model-generated supervision when candidate answers are embedded in the input.

\begin{table}[t]
\centering\footnotesize\setlength{\tabcolsep}{3pt}
\begin{tabular}{@{}l ccc c@{}}
\toprule
 & \multicolumn{3}{c}{Label \ADE $\downarrow$} & Copies B \\
\cmidrule(lr){2-4}
Label source & all & social & object & (\% rows) \\
\midrule
InternVL3-8B, leaky prompt & 0.286 & 0.149 & 0.403 & 93.4\% \\
Qwen3-VL-8B, leaky prompt & 0.249 & 0.147 & 0.336 & 81.1\% \\
Copy-B & 0.284 & 0.138 & 0.409 & 100\% \\
\midrule
InternVL3-8B, no-leak prompt & 0.334 & 0.325 & 0.342 & n/a \\
Qwen3-VL-8B, no-leak prompt & 0.247 & 0.226 & 0.264 & n/a \\
\midrule
Image centre & 0.294 & 0.284 & 0.302 & n/a \\
Gaze-LLE & 0.114 & 0.094 & 0.131 & n/a \\
MTGS & 0.099 & 0.082 & 0.116 & n/a \\
\bottomrule
\end{tabular}

\caption{\textbf{Two VLMs as labellers} (test, \R{svlm.pooled.rows} on-screen rows, three datasets). Label \ADE against the annotation. `Copies B' = share of the rows the model labelled whose label lies within 0.001 of the face centre supplied in the leaky prompt (100\% for copy-B by design; n/a = no coordinate in the prompt). Image centre, Gaze-LLE and MTGS rows are for reference.}
\label{tab:vlm}
\end{table}

\section{Checking model-generated labels before and after training}
\label{sec:check}
Label error is an inexpensive proxy that reliably separates coarse differences in student performance.
Scoring a label set against the annotation involves no student training, takes seconds and closely ranks performance (\cref{fig:labelstudent}, \cref{tab:main}).
Across the ten label sources other than the annotation (whose label error is zero by definition), Spearman's rank correlation between validation label error and student test error is \R{lcv.rho.tiny} (ViT-Tiny) and \R{lcv.rho.small} (ViT-Small).
The check also needs little annotated data: with \R{lcv.k10.clips} randomly selected validation clips (about \R{lcv.k10.frames} frames, \R{lcv.k10.trainshare} of the \R{pop.train.all} training rows), the two specialist teachers rank highest in \R{lcv.k10.top2} of \R{lcv.draws} random draws; with \R{lcv.k20.clips} clips, in \R{lcv.k20.top2}.
Training on specialist labels uses no target-dataset gaze-point annotation; selecting a strong teacher requires only a small annotated sample.

\paragraph{Example: confidence gating.}
Gating replaces low-confidence detector labels with the annotation (\cref{sec:setup}), but the label check shows that confidence does not identify incorrect labels.
When the same share of detector labels (\R{gate.rate55.all}) is replaced, confidence-guided replacement performs worse than replacing at random (object label error \R{lab.test.all.ovdgate55.object} vs \R{lab.test.all.ovdmixrand.object}; \cref{tab:main}).
Consistent with this, confidence gating is never significantly better across six student-training tests (three thresholds $\times$ two sizes), while random replacement is significantly better twice (ViT-Tiny at 0.55, \R{sig.tiny.ovdmixrand.vs.ovdgate55.all.diff}; ViT-Small at 0.40, \R{sig.small.ovdmix40.vs.ovdgate40.all.diff}; supplementary).
Replacing every detector label with the annotation makes the object labels almost perfect (object label error \R{lab.test.all.ovdgate100.object}, test rows), yet the students stay far from the annotation-trained ones (ViT-Tiny \R{st.tiny.ovdgate100.pooled} vs \R{st.tiny.gt.pooled}; ViT-Small \R{st.small.ovdgate100.pooled} vs \R{st.small.gt.pooled}): the no-leak VLM's social labels (label error \R{lab.test.all.vlmnoleak.social}), not the detector, hold these students back.

\begin{figure}[t]
\centering
\includegraphics[width=\linewidth]{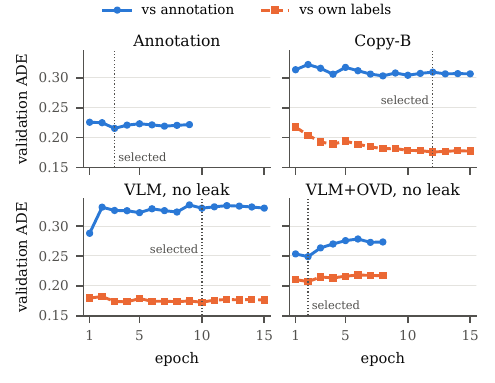}
\caption{\textbf{Validation error per epoch} (ViT-Tiny, seed 42). Panels: annotation, copy-B, no-leak VLM, VLM+OVD (no leak). Blue = against the annotation; orange = against the student's own labels (absent in the annotation panel, where both are the same); dotted line = selected epoch, chosen on own labels. Copy-B and no-leak VLM students look better than the annotation-trained student against their own labels, but not against the annotation.}
\label{fig:curves}
\end{figure}

\paragraph{Judge students against the annotation.}
When evaluated using their own labels on validation, students trained on copy-B (\R{st.tiny.copyb.val.vslabels}) and on the no-leak VLM (\R{st.tiny.vlmnoleak.val.vslabels}) appear better than the annotation-trained student (\R{st.tiny.gt.val.vslabels}).
Compared with the annotation, however, their errors rise to \R{st.tiny.copyb.val.vsgt} and \R{st.tiny.vlmnoleak.val.vsgt} (\cref{fig:curves}; ViT-Tiny).
A student that fits its teacher well can still be a poor gaze model; error should therefore be measured on annotated data, however small the sample.

\paragraph{Report spread and PCK with \ADE.}
\ADE rewards hedging towards the image centre.
The VLM+OVD students (no leak, gated, random swap) reach their best validation score at epoch 1 or 2, when their prediction spread (standard deviation of predictions over that of the annotation) has collapsed to just \R{st.tiny.ovdfamily.collapse.min}-\R{st.tiny.ovdfamily.collapse.max} of the annotation's, compared with \R{st.tiny.gt.collapse.min}-\R{st.tiny.gt.collapse.max} for the annotation-trained student.
This can produce misleading rankings: random-swap students outperform leaky-VLM students on \ADE (\R{st.tiny.ovdmixrand.pooled} vs \R{st.tiny.teacher_internvl3_alone.pooled}) yet achieve a PCK$_{.1}$ only marginally above a constant centre predictor (\R{st.tiny.ovdmixrand.pck10} vs \R{st.tiny.centre.pck10}).
We therefore recommend reporting prediction spread and PCK alongside \ADE.

\paragraph{Choose on held-out data and test by clip.}
Every student also predicts whether the target is off-screen; this head is trained on the annotated flag for every source.
At the default 0.5 threshold, F1 is only \R{io.tiny.gt.f1half} (annotation-trained ViT-Tiny); with the threshold chosen on validation, \R{io.tiny.gt.f1val} (choosing on the test rows would give \R{io.tiny.gt.f1oracle}).
Every source improves in the same way (F1 0.07-0.18 at 0.5, 0.23-0.35 with the validation threshold), because off-screen targets are rare in validation (\R{io.val.prev} of rows, against \R{io.test.prev} on test) and the head's scores rarely reach 0.5.
We evaluate temporal heads that read the previous eight frames to predict gaze 100-500\,ms ahead (supplementary) across \R{tp.tiny.runs} ViT-Tiny runs (8 sources $\times$ 3 seeds $\times$ 3 horizons; every source except the gated, random-swap and image-centre ones; \cref{tab:protocol}).
With the head architecture picked on the test rows, and rows treated as independent, the heads surpass copying the static prediction forward on object gaze in \R{tp.tiny.proto.evalpick_row.object.REALp} of \R{tp.tiny.runs} runs.
However, selecting architectures on a held-out fold and estimating uncertainty across clips reduces this count to \R{tp.tiny.proto.sel_clip.object.REALp}, of which \R{tp.tiny.proto.sel_clip.object.REALp.biased} arise from the three biased label sources (leaky VLM, copy-B, no-leak VLM).
For the leaky VLM, the apparent gain reflects static bias correction rather than anticipation, as a one-parameter correction towards the dataset mean does as well (\R{tp.tiny.teacher_internvl3_alone.h100.shrink.ade} vs \R{tp.tiny.teacher_internvl3_alone.h100.head.ade} for the head, at 100\,ms).
Even on windows where gaze changes position (\R{shift.tiny.gt.h500.shift.share} at 500\,ms), the annotation-trained student's eight-frame head performs no better than the corresponding single-frame model (\R{shift.tiny.gt.h500.shift.vsctx1.diff} \ADE, not significant).
ViT-Small shows the same pattern: heads on annotation, MTGS and Gaze-LLE students are never significantly better than copying the static prediction forward (\R{tp.small.gt.v.pooled.REALp} of 9 runs each); gains appear only for biased students (no-leak VLM, \R{tp.small.vlmnoleak.v.object.REALp} of 9 on object gaze).
Training the student and the temporal head end to end does not help either (ViT-Tiny, annotation labels, 300\,ms: \R{e2e.tiny.gt.h300.e2e.ade} vs \R{e2e.tiny.gt.h300.copy.ade} for copying forward; no seed significant).

\begin{table}[t]
\centering
\footnotesize
\setlength{\tabcolsep}{3pt}
\begin{tabular}{@{}l ccc ccc@{}}
\toprule
 & \multicolumn{3}{c}{object} & \multicolumn{3}{c}{social} \\
\cmidrule(lr){2-4}\cmidrule(l){5-7}
Protocol & better & n.s. & worse & better & n.s. & worse \\
\midrule
Picked on test rows, row CI & 50 & 9 & 13 & 29 & 1 & 42 \\
Picked on held-out fold, row CI & 43 & 14 & 15 & 24 & 9 & 39 \\
Held-out fold, clip CI (ours) & 28 & 44 & 0 & 15 & 34 & 23 \\
\bottomrule
\end{tabular}

\caption{\textbf{The same \R{tp.tiny.runs} temporal runs scored three ways} (ViT-Tiny; 8 sources $\times$ 3 seeds $\times$ 3 horizons, 100/300/500\,ms). Cells count runs where the selected temporal head is significantly better / not significantly different / significantly worse than copying the static prediction forward. Rows: head architecture picked on the test rows with row-level CIs; picked on a held-out fold with row-level CIs; held-out fold with clip-level CIs (ours).}
\label{tab:protocol}
\end{table}

\section{Limitations and conclusion}
\label{sec:conclusion}
\paragraph{Limitations.}
For every source, including Gaze-LLE, both the head boxes and the off-screen flag come from the annotation, so a deployed system would still need a head detector.
We evaluate two zero-shot VLMs (InternVL3-8B, and Qwen3-VL-8B at label level only) with two prompt designs; improved prompting (\eg a visual mark on the person~\cite{gupta2024vlmgaze}) or fine-tuning~\cite{wang2026eyesonvlm} may narrow the gap to specialist models.
Tests are performed on two student sizes using one fixed training recipe and no per-source tuning.
None of the datasets represents a robot's operational environment: all three are online videos, and no student was deployed on a robot.
MTGS was trained on VAT and ChildPlay (addressed by the VACATION check and by Gaze-LLE, \cref{sec:setup}), and DINOv2, InternVL3, Qwen3-VL and Grounding DINO were trained on web-scale data that may include frames from these public datasets.
The label check ranks coarse differences between teachers but does not separate teachers whose students lie within seed noise; ten clips identify the top two teachers but cannot determine their order.
Temporal experiments use frozen-feature heads plus one end-to-end student (ViT-Tiny, annotation labels); over 100-500\,ms, gaze moves in few windows (\R{shift.tiny.gt.h500.shift.share} at 500\,ms), which limits how much any anticipation could show.
Why Gaze-LLE's labels are easier for students to fit remains unclear.
Temporal smoothing by the teachers, clip-level annotation variability and centre bias were tested and do not account for it (supplementary material), whereas the roles of regularisation and of compatibility between label structure and student capacity remain untested.
Disentangling these mechanisms would help establish general principles for selecting and designing teachers beyond gaze estimation.

\paragraph{Conclusion.}
A specialist gaze model never trained on the target datasets' gaze annotation can replace it for training lightweight students (comparable on ViT-Tiny, significantly more accurate on ViT-Small).
VLM-generated labels can reproduce a coordinate supplied in the prompt rather than reflect gaze reasoning; running a copy-the-coordinate baseline exposes this, and with the leak removed, both VLMs remain well behind a specialist.
The best teachers can be identified before any student is trained, by measuring label error on a few annotated clips.
Students should then be evaluated against the annotation, with spread and PCK reported alongside \ADE, choices made on held-out data and differences tested by clip.
The same check can screen new teachers as they appear, and extending it to unlabelled robot video is a natural next step.

{\small
\bibliographystyle{ieeenat_fullname}
\bibliography{main}
}
\clearpage
\appendix
\setcounter{table}{0}
\setcounter{figure}{0}
\renewcommand{\thetable}{S\arabic{table}}
\renewcommand{\thefigure}{S\arabic{figure}}
\renewcommand{\theHtable}{supp.\arabic{table}}
\renewcommand{\theHfigure}{supp.\arabic{figure}}
\twocolumn[{\centering\Large\bfseries Supplementary Material\par\vspace{1em}}]

\section*{Contents}
\noindent Where to find the material the main paper refers to:
\begin{itemize}\setlength{\itemsep}{0pt}\small
  \item \textbf{VLM prompts} (leaky, no-leak, VLM+OVD) and how copy-B picks its coordinate: Sec.~\ref{sec:s-prompts}
  \item \textbf{Training recipe} of the students: Sec.~\ref{sec:s-recipe}, Tab.~\ref{tab:s-recipe}
  \item \textbf{Epoch selection on the annotation} instead of own labels: Sec.~\ref{sec:s-selection}, Tab.~\ref{tab:s-selection}
  \item \textbf{Results by dataset and gaze type}; VAT frames without person B: Sec.~\ref{sec:s-perdataset}, Tabs.~\ref{tab:s-static}-\ref{tab:s-matched}
  \item \textbf{Confidence gating vs random swap} at thresholds 0.40, 0.48, 0.55 and 1: Sec.~\ref{sec:s-gating}, Tabs.~\ref{tab:s-gaterates}-\ref{tab:s-gateepochs}
  \item \textbf{Why Gaze-LLE's labels train students as well as the annotation}: Sec.~\ref{sec:s-why}, Tabs.~\ref{tab:s-why-temporal}-\ref{tab:s-why-fit}
  \item \textbf{Second VLM} (Qwen3-VL) and copy rates per dataset: Sec.~\ref{sec:s-qwen}, Tab.~\ref{tab:s-copyrates}
  \item \textbf{Choosing a teacher from a few validation clips}; validation label errors behind $\rho=\R{lcv.rho.tiny}$: Sec.~\ref{sec:s-labelcheck}, Tabs.~\ref{tab:s-labelcheck}-\ref{tab:s-labelval}
  \item \textbf{Off-screen detection}: Sec.~\ref{sec:s-inout}, Tab.~\ref{tab:s-inout}
  \item \textbf{Temporal prediction} (ViT-Tiny and ViT-Small, controls, end-to-end): Sec.~\ref{sec:s-temporal}, Tabs.~\ref{tab:s-temporal-tiny}-\ref{tab:s-e2e}
  \item \textbf{UCO-LAEO}, a mutual-gaze-only test: Sec.~\ref{sec:s-ucolaeo}, Tab.~\ref{tab:s-ucolaeo}
  \item \textbf{Reproducibility notes}: Sec.~\ref{sec:s-repro}
\end{itemize}

\section{Label sources: prompts and parsing}
\label{sec:s-prompts}
Both VLMs (InternVL3-8B and Qwen3-VL-8B) receive the frame and a text prompt.
Decoding is greedy throughout, and Qwen3-VL-8B is run in bfloat16 precision.
The prompts below are used unchanged for both models.
Person B is the first other annotated person in the frame, in annotation-file order, and is chosen without reference to anyone's gaze.
Frames with no second person occur only in VAT; there, the leaky pipeline falls back to the annotation (\R{unmatched.share.vat} of VAT training rows).
Predictions are parsed from the first line matching \texttt{GAZE\_XY: (x, y)}.
Coordinates outside $[0,1]$ would indicate a format mismatch, but no InternVL3 no-leak prediction (about 498k) and no Qwen3-VL prediction falls outside $[0,1]$; the poor accuracy of the no-leak VLM therefore cannot be attributed to coordinate-format errors.

For object rows, VLM+OVD first asks InternVL3 to name the target with the identification prompt below.
The name is passed to Grounding DINO (base), and the centre of the highest-scoring box is used as the label.
This prompt receives the whole frame and is not told whose gaze to estimate; when several people are present, the model must infer both which person is meant and what that person is looking at.

\begin{nolinenumbers}
\medskip\noindent\textbf{Leaky prompt} (person B's head box and face centre supplied):
{\tiny
\begin{verbatim}
Gaze estimation task. Image: {W}x{H}px.

PERSON A (predict gaze): head box (normalised) (ax1,ay1) to (ax2,ay2)
PERSON B (other person): head box px (bx1,by1) to (bx2,by2), face centre (bcx,bcy)

YOUR FIRST LINE MUST BE:
GAZE_XY: (x, y)

Rules:
- If A looks at B's face -> use B's face centre: (bcx,bcy)
- If A looks at an object -> estimate object location in normalised coords
- If A looks off-screen -> use (-1,-1)
- x,y are normalised 0.0-1.0 (0,0=top-left, 1,1=bottom-right)

After the coordinate, add one sentence explaining why.
\end{verbatim}
}

\noindent The prompt is printed exactly as sent: person B's head box is given in pixels, while its face centre is normalised (the box centre divided by the image width and height); this is the coordinate copy-B returns.

\medskip\noindent\textbf{No-leak prompt} (no information about any other person):
{\tiny
\begin{verbatim}
Gaze estimation task. Image: {W}x{H}px.

PERSON A (predict gaze): head box (normalised) (ax1,ay1) to (ax2,ay2)

YOUR FIRST LINE MUST BE:
GAZE_XY: (x, y)

Rules:
- Work out where Person A is looking from their head pose and eye
  direction, then locate that point in the image yourself.
- If A is looking at another person, answer with the centre of THAT
  person's face.
- If A is looking at an object, answer with the centre of that object.
- If A is looking outside the frame, answer (-1,-1).
- x,y are normalised 0.0-1.0 (0,0=top-left, 1,1=bottom-right)

After the coordinate, add one sentence explaining why.
\end{verbatim}
}

\medskip\noindent\textbf{VLM+OVD identification prompt} (InternVL3; object rows):
{\tiny
\begin{verbatim}
Analysis Steps:
1. Examine the subject's face orientation and precise eye/pupil direction.
2. Trace the vector of their gaze to find the MOST SPECIFIC point they are
   looking at -- this could be a person's face, a person's hand or another
   specific body part, an object, or entirely outside the visible frame.
3. If they are looking at a particular body part of a person (e.g. an
   outstretched hand, not that person's face) -- identify that specific
   body part, not just the person. Only use a general person descriptor
   if no specific body part is the clear target.
4. Write 1-2 sentences of step-by-step reasoning.

If looking at a specific body part of a person (not their face/head):
Answer with the body part plus a brief visual descriptor of that person
(e.g. 'hand of man in green coat', 'outstretched arm of woman in red').
3-6 words maximum.
If looking at a person's face/head, or at the person generally with no
specific body part evident: Answer with a brief visual descriptor of that
person only (e.g. 'person in gray suit', 'woman on the right', 'man with
beard'). 2-5 words maximum.
If looking at an object: Answer with ONLY the object name plus a brief
location hint if multiple similar objects exist (e.g. 'newspaper', 'red
coffee mug', 'plate on the left'). 2-5 words maximum.

PERSPECTIVE RULE: ALL location hints (left/right/etc.) must use the
VIEWER's perspective looking AT the image -- NOT the subject's own
left/right, and NOT any other person's left/right. Worked example: if a
photo shows three people and the middle person's own right hand points
toward the person standing on the LEFT side of the photo as the viewer
sees it, the correct answer is 'person on the left' (viewer's left), even
though it is the middle person's OWN right hand doing the pointing. Always
describe positions exactly as a viewer looking at the photo would
describe them.
If looking off-screen: answer 'off-screen'.
\end{verbatim}
}
\end{nolinenumbers}

\section{Student and training recipe}
\label{sec:s-recipe}
The training configuration is identical across all label sources and both student sizes (\cref{tab:s-recipe}), with no per-source tuning.
The full image and the head crop are encoded by the same ViT backbone, and their class tokens are concatenated before prediction.
Training stops early when validation error against the source's own labels has not improved for six epochs.
\begin{table}[!htb]\centering\scriptsize\setlength{\tabcolsep}{3pt}
\begin{tabular}{@{}l p{0.70\linewidth}@{}}
\toprule
Setting & Value \\
\midrule
Backbone & ViT-Tiny/16 or ViT-Small/16 (timm, ImageNet weights), shared by both inputs \\
Inputs & scene $224{\times}224$; head crop $112{\times}112$, resized to 224 \\
Head & concat.\ class tokens $\to$ MLP (256, 128; LayerNorm, GELU, dropout 0.2/0.1) \\
Outputs & gaze point (sigmoid, 2-d); off-screen logit \\
Loss & smooth-$\ell_1$ on on-screen rows + binary cross-entropy (weight 1) \\
Optimiser & AdamW, lr $3{\times}10^{-5}$, weight decay $10^{-4}$, gradient clip 1.0 \\
Schedule & cosine to 1\% of lr over 15 epochs; batch 32 \\
Early stopping & patience 6 epochs (validation error on the source's own labels) \\
Epoch selection & lowest validation error on the source's own labels \\
Augmentation & horizontal flip (gaze mirrored), colour jitter; training only \\
Seeds & 42, 88, 137 \\
Hardware & A100 MIG slices (ViT-Tiny 1g.10gb, ViT-Small 3g.40gb) \\
\bottomrule
\end{tabular}
\caption{\textbf{Student training settings,} identical for every label source and both student sizes.}
\label{tab:s-recipe}
\end{table}

\section{Epoch selection on the annotation}
\label{sec:s-selection}
These are the same runs as in the main paper, but each source's checkpoint is chosen by validation error against the annotation instead of its own labels: an oracle choice that a user without annotation cannot make.
For the annotation, MTGS and Gaze-LLE, the selected students' test error is unchanged.
For weaker sources, annotation-based selection favours early checkpoints with collapsed predictions (\eg epoch 1 for every no-leak VLM seed).
The collapse columns give the spread of predictions relative to the annotation's spread (1 = same spread; \cref{tab:s-selection}).
\begin{table*}[!t]\centering\scriptsize\setlength{\tabcolsep}{2.5pt}
\begin{tabular}{@{}l ccc ccc@{}}
\toprule
 & \multicolumn{3}{c}{select on own labels} & \multicolumn{3}{c}{select on annotation} \\
\cmidrule(lr){2-4}\cmidrule(l){5-7}
Supervision & test ADE & epochs & collapse & test ADE & epochs & collapse \\
\midrule
\C{gt} & 0.208\,{\scriptsize$\pm$0.003} & 3, 8, 4 & 0.76-0.83 & 0.208\,{\scriptsize$\pm$0.003} & 3, 8, 4 & 0.76-0.83 \\
\C{mtgs} & 0.204\,{\scriptsize$\pm$0.004} & 13, 11, 11 & 0.82-0.84 & 0.204\,{\scriptsize$\pm$0.004} & 13, 14, 11 & 0.82-0.86 \\
\C{gazelle} & 0.204\,{\scriptsize$\pm$0.000} & 11, 11, 13 & 0.80-0.82 & 0.204\,{\scriptsize$\pm$0.000} & 11, 11, 13 & 0.80-0.82 \\
\C{teacher_hybrid} & 0.247\,{\scriptsize$\pm$0.003} & 4, 8, 3 & 0.68-0.71 & 0.246\,{\scriptsize$\pm$0.001} & 4, 9, 7 & 0.68-0.75 \\
\C{teacher_internvl3_alone} & 0.291\,{\scriptsize$\pm$0.001} & 11, 8, 10 & 0.77-0.80 & 0.291\,{\scriptsize$\pm$0.008} & 4, 11, 4 & 0.78-0.85 \\
\C{copyb} & 0.294\,{\scriptsize$\pm$0.005} & 12, 14, 10 & 0.90-0.91 & 0.291\,{\scriptsize$\pm$0.004} & 8, 13, 15 & 0.86-0.91 \\
\C{vlmnoleak} & 0.333\,{\scriptsize$\pm$0.007} & 10, 10, 1 & 0.40-0.69 & 0.308\,{\scriptsize$\pm$0.016} & 1, 1, 1 & 0.06-0.40 \\
\C{ovdalwaysdino} & 0.275\,{\scriptsize$\pm$0.004} & 2, 2, 2 & 0.23-0.37 & 0.271\,{\scriptsize$\pm$0.008} & 2, 1, 1 & 0.06-0.23 \\
\C{ovdgate55} & 0.267\,{\scriptsize$\pm$0.003} & 1, 2, 1 & 0.06-0.17 & 0.267\,{\scriptsize$\pm$0.003} & 1, 2, 1 & 0.06-0.17 \\
\C{ovdmixrand} & 0.261\,{\scriptsize$\pm$0.005} & 2, 2, 2 & 0.05-0.10 & 0.261\,{\scriptsize$\pm$0.005} & 2, 2, 2 & 0.05-0.10 \\
\C{centre} & 0.294\,{\scriptsize$\pm$0.000} & 10, 5, 15 & 0.00-0.00 & 0.294\,{\scriptsize$\pm$0.000} & 1, 1, 4 & 0.00-0.01 \\
\bottomrule
\end{tabular}

\caption{\textbf{Epoch selection on own labels vs on the annotation} (ViT-Tiny): test \ADE, selected epochs (3 seeds) and collapse (spread ratio). The image-centre row is a student trained on labels that always point to the image centre: its coordinate output is therefore constant (the constant prediction of the main paper), while its off-screen output is trained on the annotation, as for every other student.}
\label{tab:s-selection}
\end{table*}

\section{Results by dataset and gaze type}
\label{sec:s-perdataset}
\Cref{tab:s-static} reports ViT-Tiny students by dataset and gaze type (ViT-Small pooled), and \cref{tab:s-labels} gives each source's label error per dataset on the test rows.
On every dataset, the two specialists' students are as accurate as or more accurate than annotation-trained ones, and every other source trails them; the order among the weaker sources varies by dataset.
In the leaky pipeline, VAT frames with no second person use the annotation directly (\R{unmatched.share.vat} of VAT training rows).
Removing these rows and retraining the annotation and leaky VLM students changes their test error by at most 0.003, within seed noise (\cref{tab:s-matched}).
\begin{table*}[!t]\centering\scriptsize\setlength{\tabcolsep}{2.5pt}
\begin{tabular}{@{}l c c c ccc cc c c c cc@{}}
\toprule
 & Labels & & \multicolumn{7}{c}{ViT-Tiny student} & & \multicolumn{2}{c}{ViT-Small student} \\
\cmidrule(lr){2-2}\cmidrule(lr){4-10}\cmidrule(l){12-13}
 & \ADE $\downarrow$ & & \multicolumn{6}{c}{\ADE $\downarrow$} & PCK$_{.1}\uparrow$ & & \ADE $\downarrow$ & PCK$_{.1}\uparrow$ \\
\cmidrule(lr){2-2}\cmidrule(lr){4-9}\cmidrule(lr){10-10}\cmidrule(lr){12-12}\cmidrule(l){13-13}
Supervision & pooled & & pooled & VAT & ChildPlay & VACATION & social & object & pooled & & pooled & pooled \\
\midrule
\C{gt} & 0 &  & 0.208\,{\scriptsize$\pm$0.003} & 0.215 & 0.192 & 0.223 & 0.178 & 0.233 & 0.201 &  & 0.196\,{\scriptsize$\pm$0.002} & 0.243 \\
\C{mtgs} & 0.099 &  & 0.204\,{\scriptsize$\pm$0.004} & 0.208 & 0.192 & 0.216 & 0.168 & 0.234 & 0.215 &  & 0.185\,{\scriptsize$\pm$0.002} & 0.282 \\
\C{gazelle} & 0.114 &  & 0.204\,{\scriptsize$\pm$0.000} & 0.209 & 0.191 & 0.217 & 0.171 & 0.232 & 0.213 &  & 0.187\,{\scriptsize$\pm$0.004} & 0.276 \\
\C{teacher_hybrid} & 0.208 &  & 0.247\,{\scriptsize$\pm$0.003} & 0.245 & 0.254 & 0.238 & 0.209 & 0.279 & 0.131 &  & 0.238\,{\scriptsize$\pm$0.006} & 0.155 \\
\C{teacher_internvl3_alone} & 0.286 &  & 0.291\,{\scriptsize$\pm$0.001} & 0.247 & 0.353 & 0.262 & 0.168 & 0.397 & 0.163 &  & 0.289\,{\scriptsize$\pm$0.005} & 0.212 \\
\C{copyb} & 0.284 &  & 0.294\,{\scriptsize$\pm$0.005} & 0.243 & 0.358 & 0.269 & 0.163 & 0.406 & 0.199 &  & 0.287\,{\scriptsize$\pm$0.000} & 0.241 \\
\C{vlmnoleak} & 0.334 &  & 0.333\,{\scriptsize$\pm$0.007} & 0.321 & 0.344 & 0.333 & 0.287 & 0.372 & 0.052 &  & 0.330\,{\scriptsize$\pm$0.002} & 0.042 \\
\addlinespace[2pt]
\C{ovdalwaysdino} & 0.289 &  & 0.275\,{\scriptsize$\pm$0.004} & 0.286 & 0.263 & 0.276 & 0.259 & 0.289 & 0.076 &  & 0.274\,{\scriptsize$\pm$0.017} & 0.073 \\
\C{ovdgate55} & 0.222 &  & 0.267\,{\scriptsize$\pm$0.003} & 0.283 & 0.248 & 0.273 & 0.258 & 0.276 & 0.066 &  & 0.269\,{\scriptsize$\pm$0.013} & 0.075 \\
\C{ovdmixrand} & 0.214 &  & 0.261\,{\scriptsize$\pm$0.005} & 0.276 & 0.240 & 0.274 & 0.252 & 0.269 & 0.069 &  & 0.270\,{\scriptsize$\pm$0.005} & 0.085 \\
\midrule
\C{centre} & 0.294 &  & 0.294\,{\scriptsize$\pm$0.000} & 0.305 & 0.288 & 0.284 & 0.284 & 0.302 & 0.061 &  & 0.294 & 0.061 \\
\bottomrule
\end{tabular}

\caption{\textbf{Students by dataset and gaze type} (ViT-Tiny; ViT-Small pooled), with each source's pooled label \ADE on the test rows; mean of 3 seeds.}
\label{tab:s-static}
\end{table*}
\begin{table}[!htb]\centering\scriptsize\setlength{\tabcolsep}{3pt}
\begin{tabular}{@{}l ccc ccc@{}}
\toprule
 & \multicolumn{3}{c}{All on-screen rows} & \multicolumn{3}{c}{Pooled} \\
\cmidrule(lr){2-4}\cmidrule(l){5-7}
Label source & VAT & ChildPlay & VACATION & all & social & object \\
\midrule
\C{mtgs} & 0.096 & 0.098 & 0.107 & 0.099 & 0.082 & 0.116 \\
\C{gazelle} & 0.106 & 0.115 & 0.124 & 0.114 & 0.094 & 0.131 \\
\C{teacher_hybrid} & 0.247 & 0.202 & 0.153 & 0.208 & 0.151 & 0.256 \\
\C{teacher_internvl3_alone} & 0.276 & 0.325 & 0.237 & 0.286 & 0.149 & 0.403 \\
\C{copyb} & 0.273 & 0.319 & 0.244 & 0.284 & 0.138 & 0.409 \\
\C{vlmnoleak} & 0.335 & 0.323 & 0.351 & 0.334 & 0.325 & 0.342 \\
\C{ovdalwaysdino} & 0.326 & 0.243 & 0.310 & 0.289 & 0.325 & 0.259 \\
\C{ovdgate55} & 0.257 & 0.165 & 0.263 & 0.222 & 0.325 & 0.134 \\
\C{ovdmixrand} & 0.246 & 0.160 & 0.255 & 0.214 & 0.325 & 0.118 \\
\C{centre} & 0.305 & 0.288 & 0.284 & 0.294 & 0.284 & 0.302 \\
\bottomrule
\end{tabular}

\caption{\textbf{Label error per dataset and gaze type,} test rows.}
\label{tab:s-labels}
\end{table}
\begin{table}[!htb]\centering\scriptsize\setlength{\tabcolsep}{3pt}
\begin{tabular}{@{}l cc cc@{}}
\toprule
 & \multicolumn{2}{c}{pooled} & \multicolumn{2}{c}{VAT} \\
\cmidrule(lr){2-3}\cmidrule(l){4-5}
Supervision & as shipped & matched & as shipped & matched \\
\midrule
\C{gt} & 0.208\,{\scriptsize$\pm$0.003} & 0.208\,{\scriptsize$\pm$0.008} & 0.215 & 0.212 \\
\C{teacher_internvl3_alone} & 0.291\,{\scriptsize$\pm$0.001} & 0.293\,{\scriptsize$\pm$0.008} & 0.247 & 0.247 \\
\bottomrule
\end{tabular}

\caption{\textbf{VAT training rows as shipped vs matched} (frames with no person B excluded): annotation and leaky VLM students, ViT-Tiny, mean of 3 seeds.}
\label{tab:s-matched}
\end{table}

\section{Confidence gating at other thresholds}
\label{sec:s-gating}
Thresholds of 0.40, 0.48 and 0.55 replace \R{gate.rate40.all}, \R{gate.rate48.all} and \R{gate.rate55.all} of detector labels with the annotation, respectively; random swap replaces the same share at random (\cref{tab:s-gaterates}).
A threshold of 1 replaces every detector label and therefore has no random counterpart.
Across six comparisons (three thresholds and two student sizes), confidence gating is never significantly better than random replacement, while random replacement is significantly better in two cases (ViT-Tiny at 0.55, ViT-Small at 0.40).
At 0.48, one ViT-Tiny random-swap seed selects a late epoch (8), which widens its interval.
Once the students fit their labels (from epoch 2), random swap has lower validation error than confidence gating at every epoch, so the result is not an artefact of the selected epoch (\cref{tab:s-gateepochs}).
\begin{table*}[!t]\centering\scriptsize\setlength{\tabcolsep}{3pt}
\begin{tabular}{@{}c c cc c ccc c ccc@{}}
\toprule
 & & \multicolumn{2}{c}{Label \ADE, object (test)} & & \multicolumn{3}{c}{ViT-Tiny student \ADE} & & \multicolumn{3}{c}{ViT-Small student \ADE} \\
\cmidrule(lr){3-4}\cmidrule(lr){6-8}\cmidrule(l){10-12}
$t$ & replaced (train) & gated & random & & gated & random & $\Delta$ [95\% CI] & & gated & random & $\Delta$ [95\% CI] \\
\midrule
0.40 & 17\% & 0.237 & 0.227 &  & 0.269\,{\scriptsize$\pm$0.004} & 0.265\,{\scriptsize$\pm$0.003} & $-$0.004 [$-$0.009, +0.000] &  & 0.278\,{\scriptsize$\pm$0.009} & 0.272\,{\scriptsize$\pm$0.013} & $-$0.006 [$-$0.010, $-$0.002] \\
0.48 & 37\% & 0.184 & 0.174 &  & 0.269\,{\scriptsize$\pm$0.004} & 0.270\,{\scriptsize$\pm$0.010} & +0.001 [$-$0.008, +0.009] &  & 0.276\,{\scriptsize$\pm$0.015} & 0.275\,{\scriptsize$\pm$0.011} & $-$0.001 [$-$0.005, +0.004] \\
0.55 & 52\% & 0.134 & 0.118 &  & 0.267\,{\scriptsize$\pm$0.003} & 0.261\,{\scriptsize$\pm$0.005} & $-$0.006 [$-$0.009, $-$0.003] &  & 0.269\,{\scriptsize$\pm$0.013} & 0.270\,{\scriptsize$\pm$0.005} & +0.001 [$-$0.007, +0.009] \\
1 & 100\% & 0.019 & (same) &  & 0.261\,{\scriptsize$\pm$0.011} & (same) & n/a &  & 0.260\,{\scriptsize$\pm$0.006} & (same) & n/a \\
\midrule
\multicolumn{12}{@{}l}{\footnotesize No substitution (always-DINO): label 0.259; ViT-Tiny 0.275\,{\scriptsize$\pm$0.004}; ViT-Small 0.274\,{\scriptsize$\pm$0.017}.} \\
\bottomrule
\end{tabular}

\caption{\textbf{Confidence-gated vs random replacement at matched rates.} Label \ADE on test object rows, as in the main paper (replaced: share of the training detector labels replaced), and student test \ADE (mean $\pm$ sd, 3 seeds). $\Delta$ = random minus gated, so negative values favour random replacement; 95\% intervals from a paired clip bootstrap. Confidence gating is never significantly better.}
\label{tab:s-gaterates}
\end{table*}
\begin{table}[!htb]\centering\scriptsize
\begin{tabular}{@{}c ccc@{}}
\toprule
Epoch & conf.\ gate & random swap & ungated \\
\midrule
1 & 0.247 & 0.251 & 0.254 \\
2 & 0.249 & 0.244 & 0.255 \\
3 & 0.266 & 0.250 & 0.268 \\
4 & 0.268 & 0.263 & 0.279 \\
5 & 0.270 & 0.266 & 0.278 \\
6 & 0.269 & 0.263 & 0.282 \\
7 & 0.272 & 0.262 & 0.280 \\
\bottomrule
\end{tabular}

\caption{\textbf{Validation error against the annotation at each epoch} (threshold 0.55, ViT-Tiny, mean of 3 seeds; only epochs reached by every seed). From epoch 2, random swap is better than confidence gating at every epoch. Ungated: no substitution (VLM+OVD, no leak).}
\label{tab:s-gateepochs}
\end{table}

\section{Why do Gaze-LLE's labels train students as well as the annotation?}
\label{sec:s-why}
Section 4 of the main paper shows that Gaze-LLE's labels, despite a test label error of \R{lab.test.all.gazelle.all}, train students as accurate as those trained on the annotation.
We test four possible explanations on the test rows (\R{st.tiny.gt.ntest} rows from 172 clips), where a pair is two consecutive frames of the same person.
Three are not supported; one is.

\paragraph{Temporal smoothness is not supported.}
On stable pairs, where the smoothed annotation moves less than 0.01, the mean frame-to-frame step is 0.011 for the annotation, 0.025 for Gaze-LLE and 0.021 for MTGS; jitter, measured as the distance from each source's own $\pm$2-frame median, is 0.006, 0.012 and 0.010 respectively.
On shift pairs, where the smoothed annotation moves more than 0.05 (867 pairs), the mean step is 0.260 for the annotation, 0.102 for Gaze-LLE and 0.121 for MTGS (\cref{tab:s-why-temporal}).
The specialists' labels therefore fluctuate more when gaze is stable and change less when it shifts, which indicates jitter and lag rather than greater temporal consistency.
The same pattern holds on all three datasets.
One caveat is that model labels are read from a $64{\times}64$ heatmap, so they move in steps of at least 1/64 (about 0.016), which inflates their small movements.

\paragraph{Annotation variability is not supported.}
For each of 165 clips (VAT 86, ChildPlay 43, VACATION 36; the test clips with at least 20 annotated rows and 10 consecutive-frame pairs), Gaze-LLE's advantage is the annotation-trained student's error minus the Gaze-LLE-trained student's error.
After controlling for dataset, the share of gaze shifts, target motion and the share of social gaze, the partial Spearman correlation between this advantage and the clip's annotation jitter is $-$0.06 [$-$0.21, 0.10] for ViT-Tiny and 0.08 [$-$0.06, 0.23] for ViT-Small, and the advantage is flat across low-, medium- and high-jitter clips.
This analysis measures variability in the test annotation rather than in the training annotation.

\paragraph{Centre bias is not supported.}
The spatial spread of the annotation, Gaze-LLE and MTGS labels is similar (standard deviation in $x$/$y$: 0.194/0.252, 0.189/0.245 and 0.194/0.244), as is their mean distance from the image centre (0.294, 0.287 and 0.292).
The students are equally centre-biased: ViT-Tiny students trained on the annotation and on Gaze-LLE predict at mean distances of 0.212 and 0.218 from the centre.
The exception is VACATION, where Gaze-LLE's labels are more central than the annotation (0.253 vs 0.284).

\paragraph{Ease of fitting is supported.}
Gaze-LLE-trained students reproduce their own labels more closely than annotation-trained students reproduce the annotation: 0.191 vs 0.208 for ViT-Tiny and 0.172 vs 0.196 for ViT-Small, both significant (\cref{tab:s-why-fit}).
Annotation-trained students are also closer to Gaze-LLE's labels (0.200) than to the annotation itself (0.208); this is descriptive only, with no interval.
Both observations are consistent with Gaze-LLE providing a signal that is easier for a lightweight student to learn.
\begin{table*}[!t]\centering\scriptsize\setlength{\tabcolsep}{3pt}
\begin{tabular}{@{}l r ccc r ccc@{}}
\toprule
 & \multicolumn{4}{c}{Stable pairs: mean step} & \multicolumn{4}{c}{Shift pairs: mean step} \\
\cmidrule(lr){2-5}\cmidrule(l){6-9}
Test rows & $n$ & Annot. & Gaze-LLE & MTGS & $n$ & Annot. & Gaze-LLE & MTGS \\
\midrule
Pooled & 43{,}792 & 0.011 & 0.025 & 0.021 & 867 & 0.260 & 0.102 & 0.121 \\
VAT & 16{,}858 & 0.002 & 0.015 & 0.011 & 271 & 0.209 & 0.052 & 0.053 \\
ChildPlay & 16{,}951 & 0.007 & 0.022 & 0.018 & 420 & 0.259 & 0.075 & 0.110 \\
VACATION & 9{,}983 & 0.034 & 0.046 & 0.041 & 176 & 0.340 & 0.247 & 0.249 \\
\midrule
Jitter (pooled) & \multicolumn{4}{l}{annot.\ 0.006, Gaze-LLE 0.012, MTGS 0.010} & & & & \\
\bottomrule
\end{tabular}

\caption{\textbf{Temporal consistency of the annotation, Gaze-LLE and MTGS labels,} test rows. Stable pairs: smoothed annotation moves less than 0.01; shift pairs: more than 0.05. Jitter is the distance to the source's own $\pm$2-frame median.}
\label{tab:s-why-temporal}
\end{table*}
\begin{table*}[!t]\centering\scriptsize\setlength{\tabcolsep}{2.5pt}
\begin{tabular}{@{}l ccc ccc@{}}
\toprule
 & \multicolumn{3}{c}{ViT-Tiny: distance to} & \multicolumn{3}{c}{ViT-Small: distance to} \\
\cmidrule(lr){2-4}\cmidrule(l){5-7}
Student trained on & annot. & Gaze-LLE & MTGS & annot. & Gaze-LLE & MTGS \\
\midrule
Annotation & 0.208 & 0.200 & 0.196 & 0.196 & 0.191 & 0.187 \\
Gaze-LLE & 0.204 & 0.191 & 0.192 & 0.187 & 0.172 & 0.173 \\
MTGS & 0.204 & 0.194 & 0.188 & n/a & n/a & n/a \\
\midrule
\multicolumn{7}{@{}l}{Own-label fit difference [95\% CI]} \\
\quad Gaze-LLE $-$ annotation & \multicolumn{3}{c}{$-$0.017 [$-$0.025, $-$0.009]} & \multicolumn{3}{c}{$-$0.024 [$-$0.031, $-$0.016]} \\
\quad MTGS $-$ annotation & \multicolumn{3}{c}{$-$0.020 [$-$0.029, $-$0.013]} & \multicolumn{3}{c}{n/a} \\
\bottomrule
\end{tabular}

\caption{\textbf{Ease of fitting.} Mean distance from each student's predictions to each label set (test rows, 3 seeds). Bottom rows: own-label fit differences with 95\% intervals from a paired clip bootstrap.}
\label{tab:s-why-fit}
\end{table*}

\section{Second VLM}
\label{sec:s-qwen}
Qwen3-VL-8B is evaluated on the same on-screen test rows (all three datasets) with the same two prompts as InternVL3; no students are trained on its labels.
Compared with InternVL3, Qwen3-VL-8B has lower label error with both prompts (Qwen3-VL minus InternVL3, paired clip bootstrap: leaky \R{svlm.pooled.boot.leaky.diff} \R{svlm.pooled.boot.leaky.ci}; no-leak \R{svlm.pooled.boot.noleak.diff} \R{svlm.pooled.boot.noleak.ci}).
With the leaky prompt, the difference is not significant on VACATION (\R{svlm.vacation.boot.leaky.diff} \R{svlm.vacation.boot.leaky.ci}).
The copying rate is lowest on ChildPlay, where Qwen3-VL reproduces the supplied coordinate on \R{svlm.childplay.qwen3vl8b_leaky.copy} of rows.
There, its leaky labels on object rows are more accurate than copy-B's (\R{svlm.childplay.qwen3vl8b_leaky.object} vs \R{svlm.childplay.copyb.object}), which is the main reason its pooled leaky label error is below copy-B's (\R{svlm.pooled.qwen3vl8b_leaky.all} vs \R{svlm.pooled.copyb.all}; main paper, Sec.~5).
\Cref{tab:s-copyrates} gives both models' copy rates for each dataset.
\begin{table}[!htb]\centering\scriptsize
\begin{tabular}{@{}l r cc@{}}
\toprule
 & & \multicolumn{2}{c}{Copies B (leaky prompt)} \\
\cmidrule(l){3-4}
Dataset & rows & InternVL3 & Qwen3-VL \\
\midrule
VAT & \R{svlm.rows} & \R{svlm.internvl3_leaky.copy} & \R{svlm.qwen3vl8b_leaky.copy} \\
ChildPlay & \R{svlm.childplay.rows} & \R{svlm.childplay.internvl3_leaky.copy} & \R{svlm.childplay.qwen3vl8b_leaky.copy} \\
VACATION & \R{svlm.vacation.rows} & \R{svlm.vacation.internvl3_leaky.copy} & \R{svlm.vacation.qwen3vl8b_leaky.copy} \\
\midrule
All three & \R{svlm.pooled.rows} & \R{svlm.pooled.internvl3_leaky.copy} & \R{svlm.pooled.qwen3vl8b_leaky.copy} \\
\bottomrule
\end{tabular}

\caption{\textbf{Copy rates with the leaky prompt,} test rows: share of rows whose label lies within 0.001 of the supplied face centre.}
\label{tab:s-copyrates}
\end{table}

\section{Choosing a teacher from a few validation clips}
\label{sec:s-labelcheck}
The label check uses validation clips only, so no test data informs the choice of teacher.
Its population is the \R{lcv.rows} on-screen validation rows from \R{lcv.clips} clips (VAT 27, ChildPlay 23, VACATION 32).
For each sample size $k$ (3, 5, 10, 20 and 40 clips), we repeat the following \R{lcv.draws} times: draw $k$ of the \R{lcv.clips} clips at random, without replacement; compute the label error of each of the ten sources other than the annotation on those clips; and record whether MTGS and Gaze-LLE have the two lowest errors, and the Spearman correlation between the ten label errors and the students' test error (\cref{tab:s-labelcheck}).
\Cref{tab:s-labelval} gives each source's label error on all \R{lcv.clips} validation clips: the values behind $\rho = \R{lcv.rho.tiny}$ in the main paper.
\begin{table}[!htb]\centering\scriptsize\setlength{\tabcolsep}{3pt}
\begin{tabular}{@{}r r c cc@{}}
\toprule
 & & specialists & \multicolumn{2}{c}{Spearman $\rho$ vs student test \ADE} \\
\cmidrule(l){4-5}
clips & frames & top two & ViT-Tiny & ViT-Small \\
\midrule
3 & 1{,}303 & 70\% & 0.79 [0.55] & 0.81 [0.56] \\
5 & 2{,}242 & 83\% & 0.84 [0.65] & 0.83 [0.66] \\
10 & 4{,}813 & 95\% & 0.87 [0.74] & 0.87 [0.75] \\
20 & 10{,}256 & 99\% & 0.90 [0.84] & 0.89 [0.81] \\
40 & 20{,}516 & 100\% & 0.91 [0.84] & 0.90 [0.82] \\
\midrule
all 82 & 42{,}170 & yes & 0.91 & 0.90 \\
\bottomrule
\end{tabular}

\caption{\textbf{Label check on $k$ random validation clips} (\R{lcv.draws} draws per $k$). Frames: median per draw. Specialists top two: share of draws in which MTGS and Gaze-LLE have the two lowest label errors. $\rho$: median [5th percentile] Spearman correlation between the ten sources' label errors and their students' test \ADE.}
\label{tab:s-labelcheck}
\end{table}
\begin{table}[!htb]\centering\scriptsize\setlength{\tabcolsep}{3pt}
\begin{tabular}{@{}l ccc@{}}
\toprule
 & \multicolumn{2}{c}{Label \ADE} & Student test \ADE \\
\cmidrule(lr){2-3}\cmidrule(l){4-4}
Supervision & validation & test & ViT-Tiny \\
\midrule
\C{mtgs} & 0.090 & \R{lab.test.all.mtgs.all} & \R{st.tiny.mtgs.pooled} \\
\C{gazelle} & 0.103 & \R{lab.test.all.gazelle.all} & \R{st.tiny.gazelle.pooled} \\
\C{teacher_hybrid} & 0.198 & \R{lab.test.all.ovd_old.all} & \R{st.tiny.teacher_hybrid.pooled} \\
\C{teacher_internvl3_alone} & 0.311 & \R{lab.test.all.vlmleak_old.all} & \R{st.tiny.teacher_internvl3_alone.pooled} \\
\C{copyb} & 0.314 & \R{lab.test.all.copyb.all} & \R{st.tiny.copyb.pooled} \\
\C{vlmnoleak} & 0.334 & \R{lab.test.all.vlmnoleak.all} & \R{st.tiny.vlmnoleak.pooled} \\
\C{ovdalwaysdino} & 0.275 & \R{lab.test.all.ovdalwaysdino.all} & \R{st.tiny.ovdalwaysdino.pooled} \\
\C{ovdgate55} & 0.197 & \R{lab.test.all.ovdgate55.all} & \R{st.tiny.ovdgate55.pooled} \\
\C{ovdmixrand} & 0.206 & \R{lab.test.all.ovdmixrand.all} & \R{st.tiny.ovdmixrand.pooled} \\
\C{centre} & 0.264 & \R{lab.test.all.centre.all} & \R{st.tiny.centre.pooled} \\
\bottomrule
\end{tabular}

\caption{\textbf{Label error on all \R{lcv.clips} validation clips} beside the test label error (main paper, Tab.~2) and the ViT-Tiny students' test \ADE.}
\label{tab:s-labelval}
\end{table}

\section{Off-screen detection}
\label{sec:s-inout}
Every student's off-screen output is trained on the annotated flag, whatever the source of the coordinate labels.
The threshold is selected on validation by maximising F1 and then applied unchanged to the test set; we also report F1 at a fixed threshold of 0.5 and at the test-oracle optimum, an upper bound (\cref{tab:s-inout}).
Off-screen targets make up \R{io.val.prev} of validation rows and \R{io.test.prev} of test rows, and all of them come from VAT.
At 0.5, F1 is poor for every source because the head's scores rarely reach 0.5.
\begin{table*}[!t]\centering\scriptsize\setlength{\tabcolsep}{4pt}
\begin{tabular}{@{}l ccccc c c cc@{}}
\toprule
 & \multicolumn{6}{c}{ViT-Tiny} & & \multicolumn{2}{c}{ViT-Small} \\
\cmidrule(lr){2-7}\cmidrule(l){9-10}
Supervision & F1 @0.5 & F1 @val-thr. & \quad VAT only & F1 oracle & AUC & val-thr.\ (max) & & F1 @val-thr. & AUC \\
\midrule
\C{gt} & 0.13 & 0.32 & 0.37 & 0.36 & 0.73 & $2.0{\times}10^{-2}$ &  & 0.39 & 0.78 \\
\C{mtgs} & 0.18 & 0.35 & 0.39 & 0.36 & 0.73 & $3.9{\times}10^{-5}$ &  & 0.39 & 0.78 \\
\C{gazelle} & 0.15 & 0.32 & 0.34 & 0.35 & 0.71 & $4.3{\times}10^{-5}$ &  & 0.37 & 0.76 \\
\C{teacher_hybrid} & 0.11 & 0.26 & 0.28 & 0.30 & 0.69 & $3.0{\times}10^{-3}$ &  & 0.32 & 0.75 \\
\C{teacher_internvl3_alone} & 0.12 & 0.29 & 0.31 & 0.34 & 0.72 & $2.5{\times}10^{-4}$ &  & 0.35 & 0.76 \\
\C{copyb} & 0.10 & 0.23 & 0.25 & 0.33 & 0.67 & $8.4{\times}10^{-4}$ &  & 0.32 & 0.67 \\
\C{vlmnoleak} & 0.10 & 0.23 & 0.25 & 0.29 & 0.65 & $3.5{\times}10^{-2}$ &  & 0.33 & 0.74 \\
\C{ovdalwaysdino} & 0.11 & 0.27 & 0.30 & 0.31 & 0.69 & $5.2{\times}10^{-2}$ &  & 0.32 & 0.77 \\
\C{ovdgate55} & 0.11 & 0.28 & 0.31 & 0.32 & 0.71 & $1.1{\times}10^{-2}$ &  & 0.33 & 0.76 \\
\C{ovdmixrand} & 0.07 & 0.25 & 0.31 & 0.29 & 0.66 & $2.7{\times}10^{-3}$ &  & 0.35 & 0.77 \\
\C{centre} & 0.12 & 0.29 & 0.31 & 0.34 & 0.73 & $1.3{\times}10^{-3}$ &  & 0.28 & 0.74 \\
\bottomrule
\end{tabular}

\caption{\textbf{Off-screen detection.} F1 at a fixed threshold of 0.5, at the threshold selected on validation (all test rows, and VAT rows only, where every positive is) and at the test-oracle optimum; AUC is threshold-independent. Val-thr.\ (max): the largest of the three seeds' selected thresholds. Mean of 3 seeds.}
\label{tab:s-inout}
\end{table*}

\section{Temporal prediction}
\label{sec:s-temporal}
Each frame is represented by a frozen static student's 128-dimensional embedding and predicted gaze point.
A temporal head receives eight consecutive frames (about 270-330\,ms) and predicts the annotated gaze point 100, 300 or 500\,ms after the last frame.
Windows are taken from the test clips only, with a stride of four frames (\R{tp.h100.rows} windows from \R{tp.h100.clips} clips at 100\,ms), because on the training clips the static students' predictions are fitted to frames they were trained on.
The temporal manifests contain 41 VACATION test clips, against 38 in the static evaluation (VAT and ChildPlay match), so the temporal analysis covers \R{tp.h100.clips} clips rather than 172.
We evaluate four temporal architectures: a GRU, a Transformer, and residual versions of each, which predict a correction to the last frame's point.
All are trained with a smooth-$\ell_1$ loss, AdamW (learning rate $10^{-5}$), batch size 64 and a cosine schedule for up to 40 epochs.

Performance is estimated with five-fold cross-fitting over clips, stratified by dataset, using the same split for every run: three folds train the head, a fourth selects the epoch and architecture, and the fifth is predicted.
Every window therefore receives one out-of-fold prediction.
We compare against three baselines: copy-forward, which repeats the static prediction for the last frame and is deployable; the last-known annotation, an oracle that needs the annotation at test time; and the global mean.
For each window, we take the head's error minus the copy-forward error; 95\% intervals come from 10,000 resamples of clips.

Three controls test where any gain comes from.
Shrink-to-mean moves the copy-forward point $c$ towards the dataset mean $\mu$, $c + a(\mu - c)$, where the shrink factor $a$ is a single number fitted on the other folds; it uses no temporal input.
The single-frame head uses the same architectures but sees only the last frame.
The end-to-end model trains the student and the head together (ViT-Tiny, annotation labels).

On gaze-shift windows, where the annotated target moves more than 0.1, the heads beat copy-forward, but on the far more common stay windows they are no better, and slightly worse on average (\cref{tab:s-shift}).
Because the eight-frame head matches the single-frame head on shift windows, the gain comes from the last frame, not from motion.
\Cref{tab:s-temporal-tiny} gives the results of the \R{tp.tiny.runs} ViT-Tiny runs for each source.
On ViT-Small, heads on annotation, MTGS and Gaze-LLE students are never significantly better than copy-forward (\cref{tab:s-temporal-small}), and end-to-end training gives no gain either (\cref{tab:s-e2e}).
For the leaky VLM, the main paper's explanation of the gain as a correction towards the mean holds on ViT-Tiny, where shrink-to-mean matches the head (\R{tp.tiny.teacher_internvl3_alone.h100.shrink.ade} vs \R{tp.tiny.teacher_internvl3_alone.h100.head.ade} at 100\,ms), but not on ViT-Small, where the head is better (\R{tp.small.teacher_internvl3_alone.h100.head.ade} vs \R{tp.small.teacher_internvl3_alone.h100.shrink.ade}).
\begin{table*}[!t]\centering\scriptsize\setlength{\tabcolsep}{2.5pt}
\begin{tabular}{@{}l cccc cc ccc@{}}
\toprule
 & \multicolumn{4}{c}{ADE at 100\,ms} & \multicolumn{2}{c}{$\Delta$ head$-$copy} & \multicolumn{3}{c}{object / social verdicts (of 9)} \\
\cmidrule(lr){2-5}\cmidrule(lr){6-7}\cmidrule(l){8-10}
Static student & copy-fwd & head & 1-frame & shrink & 300\,ms & 500\,ms & better & n.s. & worse \\
\midrule
\C{gt} & 0.205 & 0.207 & 0.209 & 0.209 & +0.002 & +0.002 & 0\,/\,0 & 9\,/\,8 & 0\,/\,1 \\
\C{mtgs} & 0.201 & 0.204 & 0.202 & 0.203 & +0.001 & +0.001 & 1\,/\,0 & 8\,/\,6 & 0\,/\,3 \\
\C{gazelle} & 0.201 & 0.205 & 0.205 & 0.204 & +0.003 & +0.003 & 0\,/\,0 & 9\,/\,8 & 0\,/\,1 \\
\C{teacher_hybrid} & 0.242 & 0.237 & 0.239 & 0.244 & $-$0.005 & $-$0.004 & 0\,/\,0 & 9\,/\,9 & 0\,/\,0 \\
\C{teacher_internvl3_alone} & 0.283 & 0.270 & 0.270 & 0.269 & $-$0.010 & $-$0.011 & 9\,/\,0 & 0\,/\,0 & 0\,/\,9 \\
\C{copyb} & 0.284 & 0.266 & 0.264 & 0.267 & $-$0.020 & $-$0.021 & 9\,/\,0 & 0\,/\,0 & 0\,/\,9 \\
\C{vlmnoleak} & 0.331 & 0.279 & 0.280 & 0.291 & $-$0.051 & $-$0.050 & 9\,/\,9 & 0\,/\,0 & 0\,/\,0 \\
\C{ovdalwaysdino} & 0.274 & 0.267 & 0.272 & 0.278 & $-$0.006 & $-$0.005 & 0\,/\,6 & 9\,/\,3 & 0\,/\,0 \\
\midrule
Global mean & 0.291 & & & & & & & & \\
Last-known annotation (oracle) & 0.012 & & & & & & & & \\
\bottomrule
\end{tabular}

\caption{\textbf{Temporal heads on ViT-Tiny students} (\R{tp.tiny.runs} runs: 8 sources $\times$ 3 seeds $\times$ 3 horizons). 1-frame: the same heads given only the last frame; shrink: copy-forward moved towards the dataset mean. $\Delta$ = head minus copy-forward \ADE; verdicts from a paired clip bootstrap.}
\label{tab:s-temporal-tiny}
\end{table*}
\begin{table*}[!t]\centering\scriptsize\setlength{\tabcolsep}{2.5pt}
\begin{tabular}{@{}l ccc ccc ccc@{}}
\toprule
 & \multicolumn{3}{c}{ADE at 100\,ms} & \multicolumn{3}{c}{$\Delta$ head$-$copy} & \multicolumn{3}{c}{object / social verdicts (of 9)} \\
\cmidrule(lr){2-4}\cmidrule(lr){5-7}\cmidrule(l){8-10}
Static student & copy-fwd & head & shrink & 100\,ms & 300\,ms & 500\,ms & better & n.s. & worse \\
\midrule
\C{gt} & 0.193 & 0.197 & 0.194 & +0.004 & +0.003 & +0.003 & 0\,/\,0 & 6\,/\,7 & 3\,/\,2 \\
\C{mtgs} & 0.183 & 0.185 & 0.184 & +0.002 & +0.002 & +0.002 & 0\,/\,0 & 9\,/\,0 & 0\,/\,9 \\
\C{gazelle} & 0.184 & 0.187 & 0.184 & +0.003 & +0.005 & +0.005 & 0\,/\,0 & 7\,/\,8 & 2\,/\,1 \\
\C{teacher_hybrid} & 0.233 & 0.230 & 0.234 & $-$0.004 & $-$0.003 & $-$0.003 & 2\,/\,0 & 7\,/\,9 & 0\,/\,0 \\
\C{teacher_internvl3_alone} & 0.280 & 0.255 & 0.266 & $-$0.025 & $-$0.023 & $-$0.020 & 9\,/\,0 & 0\,/\,0 & 0\,/\,9 \\
\C{copyb} & 0.277 & 0.259 & 0.264 & $-$0.018 & $-$0.020 & $-$0.019 & 9\,/\,0 & 0\,/\,0 & 0\,/\,9 \\
\C{vlmnoleak} & 0.330 & 0.276 & 0.291 & $-$0.055 & $-$0.052 & $-$0.050 & 9\,/\,9 & 0\,/\,0 & 0\,/\,0 \\
\C{ovdalwaysdino} & 0.274 & 0.251 & 0.272 & $-$0.023 & $-$0.018 & $-$0.018 & 1\,/\,6 & 8\,/\,3 & 0\,/\,0 \\
\bottomrule
\end{tabular}

\caption{\textbf{Temporal heads on ViT-Small students.} As with ViT-Tiny, annotation-, MTGS- and Gaze-LLE-trained students are never significantly better than copying the static prediction forward; gains appear only for the biased label sources. $\Delta$ = head minus copy-forward \ADE; verdicts from a paired clip bootstrap (3 seeds $\times$ 3 horizons).}
\label{tab:s-temporal-small}
\end{table*}
\begin{table*}[!t]\centering\scriptsize\setlength{\tabcolsep}{3pt}
\begin{tabular}{@{}l c rr ccc@{}}
\toprule
 & & \multicolumn{2}{c}{windows} & \multicolumn{3}{c}{8-frame head minus \dots{} [95\% CI]} \\
\cmidrule(lr){3-4}\cmidrule(l){5-7}
Student & $h$ & type & share & copy-forward & shrink-to-mean & 1-frame head \\
\midrule
\C{gt} & 300\,ms & shift & 8\% & $-$0.008 [$-$0.012, $-$0.004] & $-$0.003 [$-$0.007, +0.000] & +0.000 [$-$0.001, +0.002] \\
\C{gt} & 300\,ms & stay & 90\% & +0.003 [$-$0.001, +0.006] & +0.003 [+0.000, +0.006] & $-$0.000 [$-$0.001, +0.001] \\
\C{gt} & 500\,ms & shift & 12\% & $-$0.010 [$-$0.014, $-$0.005] & $-$0.004 [$-$0.008, $-$0.000] & +0.001 [$-$0.001, +0.002] \\
\C{gt} & 500\,ms & stay & 85\% & +0.003 [$-$0.001, +0.007] & +0.003 [+0.000, +0.007] & +0.000 [$-$0.001, +0.001] \\
\midrule
\C{mtgs} & 300\,ms & shift & 8\% & $-$0.009 [$-$0.013, $-$0.005] & $-$0.004 [$-$0.007, $-$0.000] & +0.002 [+0.000, +0.003] \\
\C{mtgs} & 300\,ms & stay & 90\% & +0.002 [$-$0.002, +0.006] & +0.002 [$-$0.001, +0.006] & $-$0.000 [$-$0.001, +0.001] \\
\C{mtgs} & 500\,ms & shift & 12\% & $-$0.011 [$-$0.014, $-$0.007] & $-$0.004 [$-$0.007, $-$0.002] & +0.001 [$-$0.000, +0.002] \\
\C{mtgs} & 500\,ms & stay & 85\% & +0.002 [$-$0.002, +0.006] & +0.002 [$-$0.000, +0.006] & $-$0.000 [$-$0.002, +0.001] \\
\midrule
\C{gazelle} & 300\,ms & shift & 8\% & $-$0.007 [$-$0.010, $-$0.004] & $-$0.005 [$-$0.008, $-$0.002] & +0.001 [$-$0.001, +0.003] \\
\C{gazelle} & 300\,ms & stay & 90\% & +0.004 [$-$0.001, +0.010] & +0.004 [$-$0.000, +0.008] & $-$0.001 [$-$0.002, +0.001] \\
\C{gazelle} & 500\,ms & shift & 12\% & $-$0.010 [$-$0.013, $-$0.006] & $-$0.007 [$-$0.010, $-$0.003] & +0.001 [$-$0.001, +0.002] \\
\C{gazelle} & 500\,ms & stay & 85\% & +0.005 [$-$0.002, +0.012] & +0.004 [$-$0.001, +0.010] & $-$0.001 [$-$0.002, $-$0.000] \\
\bottomrule
\end{tabular}

\caption{\textbf{Shift versus stay windows} (ViT-Tiny, 3 seeds). A shift window is one where the annotated target moves more than 0.1 between the last context frame and the horizon. The eight-frame head beats copy-forward on shift windows but not on stay windows, and matches the single-frame head on both. Values: eight-frame head minus each comparison, with 95\% intervals from a paired clip bootstrap.}
\label{tab:s-shift}
\end{table*}
\begin{table*}[!t]\centering\scriptsize\setlength{\tabcolsep}{2.5pt}
\begin{tabular}{@{}l c ccc c ccc@{}}
\toprule
 & & \multicolumn{3}{c}{ADE (matched rows)} & & \multicolumn{3}{c}{e2e vs copy-fwd: pooled / object / social} \\
\cmidrule(lr){3-5}\cmidrule(l){7-9}
Horizon & seeds & copy-fwd & frozen head & end-to-end & oracle & better & n.s. & worse \\
\midrule
300\,ms & 3 & 0.208 & 0.210 & 0.208 & 0.035 & 0\,/\,1\,/\,0 & 3\,/\,2\,/\,2 & 0\,/\,0\,/\,1 \\
500\,ms & 1 & 0.212 & 0.215 & 0.212 & 0.053 & 0\,/\,0\,/\,0 & 1\,/\,1\,/\,1 & 0\,/\,0\,/\,0 \\
\bottomrule
\end{tabular}

\caption{\textbf{End-to-end training} (ViT-Tiny, annotation labels). Training the student and the temporal head together does not beat copy-forward or the frozen-feature head, so frozen features are not what limits temporal prediction here. \ADE on matched rows; oracle = last-known annotation.}
\label{tab:s-e2e}
\end{table*}

\section{UCO-LAEO: a mutual-gaze-only test}
\label{sec:s-ucolaeo}
UCO-LAEO is used only for evaluation; no student is trained on it.
It contains \R{uco.rows} rows, all showing mutual gaze between two people.
MTGS was trained on UCO-LAEO as part of its original training data.
Because every row contains mutual gaze, the other person's face centre is almost always the correct target: copy-B's labels are nearly identical to the annotation (label error 0.001).
As a result, copy-B gives the best students (ViT-Small \ADE \R{uco.small.copyb.ade}), and students trained on leaky VLM labels outperform annotation-trained students (\R{uco.small.teacher_internvl3_alone.ade} vs \R{uco.small.gt.ade}; \cref{tab:s-ucolaeo}).
We therefore exclude UCO-LAEO from the main results, as its mutual-gaze-only structure rewards the shortcut under investigation rather than general gaze-target estimation.
\begin{table*}[!t]\centering\scriptsize\setlength{\tabcolsep}{3pt}
\begin{tabular}{@{}l c c cc c cc@{}}
\toprule
 & Labels & & \multicolumn{2}{c}{ViT-Tiny} & & \multicolumn{2}{c}{ViT-Small} \\
\cmidrule(lr){2-2}\cmidrule(lr){4-5}\cmidrule(l){7-8}
Supervision & \ADE & & \ADE & PCK$_{.1}$ & & \ADE & PCK$_{.1}$ \\
\midrule
\C{gt} & 0 &  & 0.192\,{\scriptsize$\pm$0.004} & 0.247 &  & 0.165\,{\scriptsize$\pm$0.005} & 0.301 \\
\C{mtgs} & 0.031 &  & 0.174\,{\scriptsize$\pm$0.001} & 0.294 &  & 0.147\,{\scriptsize$\pm$0.003} & 0.399 \\
\C{gazelle} & 0.059 &  & 0.176\,{\scriptsize$\pm$0.004} & 0.278 &  & 0.152\,{\scriptsize$\pm$0.007} & 0.348 \\
\C{teacher_hybrid} & 0.030 &  & 0.204\,{\scriptsize$\pm$0.015} & 0.190 &  & 0.168\,{\scriptsize$\pm$0.008} & 0.302 \\
\C{teacher_internvl3_alone} & 0.030 &  & 0.159\,{\scriptsize$\pm$0.001} & 0.343 &  & 0.138\,{\scriptsize$\pm$0.010} & 0.423 \\
\C{copyb} & 0.001 &  & 0.145\,{\scriptsize$\pm$0.008} & 0.415 &  & 0.109\,{\scriptsize$\pm$0.006} & 0.618 \\
\C{vlmnoleak} & 0.322 &  & 0.313\,{\scriptsize$\pm$0.024} & 0.042 &  & 0.321\,{\scriptsize$\pm$0.007} & 0.042 \\
\C{ovdalwaysdino} & 0.322 &  & 0.268\,{\scriptsize$\pm$0.004} & 0.052 &  & 0.277\,{\scriptsize$\pm$0.032} & 0.047 \\
\C{ovdgate55} & 0.322 &  & 0.263\,{\scriptsize$\pm$0.001} & 0.060 &  & 0.290\,{\scriptsize$\pm$0.035} & 0.049 \\
\C{ovdmixrand} & 0.322 &  & 0.262\,{\scriptsize$\pm$0.004} & 0.055 &  & 0.312\,{\scriptsize$\pm$0.008} & 0.038 \\
\C{centre} & 0.283 &  & 0.283\,{\scriptsize$\pm$0.000} & 0.028 &  & 0.283\,{\scriptsize$\pm$0.000} & 0.028 \\
\bottomrule
\end{tabular}

\caption{\textbf{Results on UCO-LAEO} (never used for training). Label \ADE against the annotation; student \ADE (mean $\pm$ sd over 3 seeds) and PCK$_{.1}$. Because every row shows mutual gaze, predicting the other person's face centre is usually correct, so copy-B gives the strongest students. Every row is social, so VLM+OVD labels equal the VLM labels with the same prompt.}
\label{tab:s-ucolaeo}
\end{table*}

\section{Reproducibility notes}
\label{sec:s-repro}
Repeated runs are bit-identical when executed on the same GPU partition type; changing partition type introduces variability comparable to that observed across random seeds.
Most ViT-Tiny students are trained on \texttt{1g.10gb} A100 partitions; four runs (two Gaze-LLE seeds and two threshold-1 runs) use \texttt{3g.40gb} partitions.
Labels are generated on a mixture of A100 partitions, full A100 GPUs and L40S GPUs; repeated greedy VLM inference can produce different outputs for a small percentage of rows.
Code will be made publicly available upon acceptance.

\end{document}